\documentclass[journal,twoside,web]{ieeecolor}
\usepackage{generic}
\usepackage{siunitx}
\usepackage{cite}
\usepackage{amsmath,amssymb,amsfonts}
\usepackage{algorithmic}
\usepackage{graphicx}
\usepackage{algorithm,algorithmic}
\usepackage{hyperref}
\hypersetup{hidelinks=true}
\usepackage{textcomp}
\usepackage{bm}
\usepackage{cuted}

\usepackage{pgfplots}
\pgfplotsset{compat=1.18}

\usepackage[toc,acronym,nomain,nonumberlist,nogroupskip,nopostdot]{glossaries}  

\usepackage{booktabs, arydshln}
\makeatletter
\def\adl@drawiv#1#2#3{%
        \hskip.5\tabcolsep
        \xleaders#3{#2.5\@tempdimb #1{1}#2.5\@tempdimb}%
                #2\z@ plus1fil minus1fil\relax
        \hskip.5\tabcolsep}
\newcommand{\cdashlinelr}[1]{%
  \noalign{\vskip\aboverulesep
           \global\let\@dashdrawstore\adl@draw
           \global\let\adl@draw\adl@drawiv}
  \cdashline{#1}
  \noalign{\global\let\adl@draw\@dashdrawstore
           \vskip\belowrulesep}}
\makeatother

\usepackage{mathtools}
\DeclareMathSymbol{\shortminus}{\mathbin}{AMSa}{"39}
\usepackage{bbm} 
\usepackage{tikz}
\usetikzlibrary{calc}

\newacronym{GSP}{GSP}{graph signal processing}
\newacronym{GSO}{GSO}{graph shift operator}
\newacronym{GFT}{GFT}{graph Fourier transform}
\newacronym{GIR}{GIR}{graph impulse response}
\newacronym{GFR}{GFR}{graph frequency response}
\newacronym{DFT}{DFT}{discrete Fourier transform}
\newacronym{TV}{TV}{total variation}
\newacronym{DSP}{DSP}{discrete Signal Processing}
\newacronym{EEG}{EEG}{electroencephalography}
\newacronym{EOG}{EOG}{electrooculography}
\newacronym{ECoG}{ECoG}{electrocorticography}
\newacronym{EMG}{EMG}{electromyography}
\newacronym{MEG}{MEG}{magnetoencephalography}
\newacronym{BCI}{BCI}{brain-computer interface}
\newacronym{fMRI}{fMRI}{functional magnetic resonance imaging}
\newacronym{BOLD}{BOLD}{blood-oxygenation-level–dependent}
\newacronym{MRI}{MRI}{magnetic resonance imaging}
\newacronym{ERP}{ERP}{event-related potential}

\newacronym{PCA}{PCA}{principal component analysis}
\newacronym{SVM}{SVM}{support vector machine}
\newacronym[longplural={power spectral densities}]{PSD}{PSD}{power spectral density}
\newacronym{PSD-N}{PSD-N}{power spectral density network}
\newacronym{CNN}{CNN}{convolutional neural network}
\newacronym{GNN}{GNN}{graph neural network}
\newacronym{GAT}{GAT}{graph attention network}
\newacronym{RNN}{RNN}{recurrent neural network}
\newacronym{LSTM}{LSTM}{long short-term memory}
\newacronym{GRU}{GRU}{gated recurrent unit}
\newacronym{MSA-CNN}{MSA-CNN}{multi-scale attention convolutional neural network}
\newacronym{MSM}{MSM}{multi-scale module}
\newacronym{TCM}{TCM}{temporal context module}
\newacronym{GCN}{GCN}{graph convolutional network}
\newacronym{ANOVA}{ANOVA}{analysis of variance}
\newacronym{SOTA}{SOTA}{state of the art}

\newacronym{ML}{ML}{machine learning}
\newacronym{MFLOPS}{MFLOPS}{Mega floating point operations per second}

\newacronym{AD}{AD}{Alzheimer's disease}
\newacronym{PD}{PD}{Parkinson's disease}
\newacronym{MMSE}{MMSE}{mini–mental state examination}
\newacronym{HC}{HC}{healthy control}
\newacronym{FTD}{FTD}{frontotemporal dementia}

\newacronym{PLI}{PLI}{phase lag index}
\newacronym{wPLI}{wPLI}{weighted phase lag index}
\newacronym{PLV}{PLV}{phase locking value}
\newacronym{Coh}{Coh}{coherence}
\newacronym{iCoh}{iCoh}{imaginary coherence}
\newacronym{AEC}{AEC}{amplitude envelope correlation}
\newacronym{MI}{MI}{mutual information}
\newacronym{VAR}{VAR}{vector autoregression}

\newacronym{ROI}{ROI}{region of interest}
\newacronym{MPNN}{MPNN}{message passing neural network}
\newacronym{TR}{TR}{repetition time}

\definecolor{dodgerblue}{rgb}{0.12, 0.56, 1.0}
\definecolor{newred}{rgb}{0.9, 0.09, 0.1}

\def\BibTeX{{\rm B\kern-.05em{\sc i\kern-.025em b}\kern-.08em
    T\kern-.1667em\lower.7ex\hbox{E}\kern-.125emX}}
\begin{document}
\title{Connectivity Estimation using Stochastic Graph Heat Modelling}
\author{Stephan Goerttler, Min Wu, \IEEEmembership{Senior Member, IEEE}, and Fei He, \IEEEmembership{Senior Member, IEEE}
\thanks{Stephan Goerttler was supported by A*STAR ARAP Scholarship. Fei He was supported by EPSRC grant [EP/X020193/1].}
\thanks{Stephan Goerttler and Fei He are with the Centre for Computational Science and Mathematical Modelling, Coventry University, Coventry CV1 5FB, UK (e-mail: goerttlers@uni.coventry.ac.uk; fei.he@coventry.ac.uk).}
\thanks{Min Wu is with the Institute for Infocomm Research, A*STAR, Singapore (e-mail: Wu\_Min@a-star.edu.sg).}
\thanks{This article has supplementary material provided by the authors.}
}

\maketitle
\markboth{Stephan Goerttler \MakeLowercase{\textit{et al.}}: Connectivity Estimation using Stochastic Graph Heat Modelling}{}

\begin{abstract}
A growing number of techniques leverage the spatial structures that underlie many real-world datasets. 
Despite these advances, the complementary task of estimating spatial structures and understanding their role within these techniques has often been overlooked.
In neurophysiological data analysis specifically, numerous methods exist to estimate brain connectivity, but most are not explicitly model-based, dynamic, multivariate, or directed.
%
To address these limitations, we previously introduced noise-driven heat modelling on graphs for neurophysiological connectivity estimation.
In this study, we extend this framework by relaxing earlier noise assumptions and adding regularisation to improve robustness.
We also develop a simulation procedure to characterise and evaluate our technique in a controlled setting. Finally, we demonstrate that the technique is able to capture meaningful spatial structure across two experiments, each using two real-world  datasets.
The explicit model formulation of our connectivity estimator has the potential to improve the interpretability of graph-based techniques across a wide range of applications. The code implementing our method is available at \url{https://github.com/sgoerttler/Heat_Connectivity}.


\end{abstract}


\section{Introduction}

\IEEEPARstart{M}{any} real-world datasets, such as social networks, transportation systems, or brain connectivity networks, reside on irregular, non-Euclidean structures, which are typically represented using graph theory \cite{wu2020comprehensive}.
Detailed knowledge and information processing on these graph structures can be exploited through emerging graph-based techniques,
such as 
\gls{GNN} and \gls{GSP}.
While much research has focused on improving model architectures and learning strategies, the choice of graph prior is equally important \cite{thanou2017learning}.
This issue is particularly relevant in neurophysiology, where graphs can be derived from a wide range of structural and functional connectivity estimation methods \cite{goerttler2024understanding}.
Given this diversity, graph priors are commonly selected based on empirical performance, interpretability, or combinations of multiple estimation methods \cite{klepl2024graph}.
However, the relationship between the estimated connectivity structures and the neurophysiological signals is often unclear. 
For example, many connectivity estimation methods, such as \gls{PLI} \cite{stam2007phase}, \gls{PLV} \cite{lachaux1999measuring}, and \gls{AEC} \cite{brookes2011investigating}, are heuristic in nature and not explicitly derived from an underlying mechanistic model of neural dynamics.
Some methods, such as pairwise Pearson correlation or covariance, as well as geometry-based structural connectivity measures, are instantaneous (same-time) metrics, which may be insufficient to capture or predict temporal dynamics. 
In addition, many methods estimate connectivity using bivariate pairwise calculations between time series, which fail to capture the multivariate, distributed dynamics of the network \cite{bastos2016tutorial}.
Lastly, many connectivity estimation methods are symmetric and fail to capture directed brain activity patterns, despite their importance for understanding neurological disorders, such as \gls{PD} \cite{mijalkov2022directed}.

\begin{figure}[!t]
    \centering
    \vspace{2pt}
    \resizebox{0.8\textwidth}{!}{%
    \renewcommand*\familydefault{\sfdefault} 
\fontfamily{\sfdefault}\selectfont

\noindent
\begin{tikzpicture}[bend angle=10]
    \clip (0.4, -2.2) rectangle (12, 1.6);
  \node[draw, circle, fill=dodgerblue!20, thick, minimum size=0.6cm] (t0) at (1.5,0){$1$};
  \node[draw, circle, fill=newred!15, thick, minimum size=0.6cm,shift={(1.4, 1.19)}] (t1) at (t0){$2$};
  \node[draw, circle, fill=dodgerblue!5, thick, minimum size=0.6cm,shift={(2.7, -1.6)}] (t2) at (t0){$6$};
  \node[draw, circle, fill=newred!100, thick, minimum size=0.6cm,shift={(2.4, 0.4)}] (t3) at (t0){$3$};
  \node[draw, circle, fill=newred!30, thick, minimum size=0.6cm,shift={(4.5, 0.8)}] (t4) at (t0){$4$};
  \node[draw, circle, fill=newred!70, thick, minimum size=0.6cm,shift={(4.8,-0.6)}] (t5) at (t0){$5$};

  \draw[-,thick,line width=0.5mm] (t1) -- (t0) node[text width=1.1cm,align=center, font={\footnotesize}, pos=0.65,above] {$L_{12\phantom{00}}$};
  \draw[-,thick,line width=0.35mm] (t0) -- (t2) node[text width=1.1cm,align=center, font={\footnotesize}, pos=0.5,below] {$L_{16}$};
  \draw[-,thick,line width=0.4mm] (t3) -- (t0) node[text width=1.1cm,align=center, font={\footnotesize}, pos=0.55,above] {$L_{13}$};
  \draw[-,thick,line width=0.2mm] (t3) -- (t4) node[text width=1.1cm,align=center, font={\footnotesize}, pos=0.2,above, yshift=-0.5pt] {$L_{34}$};
  \draw[-,thick,line width=0.2mm] (t1) -- (t4) node[text width=1.1cm,align=center, font={\footnotesize}, pos=0.4,above] {$L_{24}$};
  \draw[-,thick,line width=0.08mm] (t2) -- (t3) node[text width=1.1cm,align=center, font={\footnotesize}, pos=0.7,below] {\phantom{0000}$L_{36}$};
  \draw[-,thick,line width=0.45mm] (t5) -- (t3) node[text width=1.1cm,align=center, font={\footnotesize}, pos=0.15,above] {$L_{35}$};%
  \draw[-,thick,line width=0.10mm] (t2) -- (t5) node[text width=1.1cm,align=center, font={\footnotesize}, pos=0.6,below] {$L_{56}$};%
  \draw[-,thick,line width=0.25mm] (t4) -- (t5) node[text width=1.1cm,align=center, font={\footnotesize}, pos=0.15,below] {\phantom{0000}\,$L_{45}$};%
  \draw[-,thick,line width=0.12mm] (t2) -- (t4) node[text width=1.1cm,align=center, font={\footnotesize}, pos=0.4,below] {\phantom{0}\,\,$L_{46}$};%



    \foreach \from/\to/\bend/\thick/\startcor in {t3/t0/15/0.8/0.25, t3/t2/-15/0.35/0.25, t3/t4/-15/0.25/0.25, t3/t5/-15/0.05/0.25} {
  \coordinate (p30) at ($(\from)!\startcor!(\to)$);
  \coordinate (p70) at ($(\from)!0.70!(\to)$);
  \path let
    \p1 = ($(\from)-(\to)$),
    \n1 = {veclen(\x1,\y1)},
    \n2 = {-6*(\bend/15)/\n1*\y1},  
    \n3 = {6*(\bend/15)/\n1*\x1}    
  in
    coordinate (p30down) at ($(p30)-(\n2,\n3)$)
    coordinate (p70down) at ($(p70)-(\n2,\n3)$);
  \draw[line width=\thick mm,bend left=\bend,newred!100] (p30down) edge[-stealth]  
    node[text width=1.1cm,align=center, font={\tiny}, pos=0.5, below,sloped] {}(p70down);
}




\foreach \x/\y/\xshift/\yshift/\xshiftarrowa/\xshiftarrowb/\yshiftarrowa/\yshiftarrowb/\xshiftnoise/\yshiftnoise in {1.5/0/-0.9/0/0.2/0.475/0/0/0/0, 2.9/1.1/-0.9/0/0.2/0.475/0/0/0/0, 4.2/-1.6/-0.9/0/0.2/0.475/0/0/-0.4/-0.35, 3.9/0.5/-0.9/0/0.2/0.475/-0.0/-0.0/0.325/-0.13, 6/0.8/0.9/0/-0.2/-0.475/0/0/0/0, 6.3/-0.6/0.9/0/-0.2/-0.475/0/0/0/0} {
  \pgfmathsetmacro{\xpos}{\x+\xshift}
  \pgfmathsetmacro{\ypos}{\y+\yshift}
  \pgfmathsetmacro{\xscale}{0.5}
  \draw[thin, gray]
    plot[smooth] coordinates {
      ({\xpos - 0.2 * \xscale}, {\ypos + .0228186358})
      ({\xpos - 0.18 * \xscale}, {\ypos - .024652121})
      ({\xpos - 0.16 * \xscale}, {\ypos + .059664526})
      ({\xpos - 0.14 * \xscale}, {\ypos + .048429195})
      ({\xpos - 0.12 * \xscale}, {\ypos + .019574632})
      ({\xpos - 0.1 * \xscale}, {\ypos - .093158828})
      ({\xpos - 0.08 * \xscale}, {\ypos + .16616722})
      ({\xpos - 0.06 * \xscale}, {\ypos + .109571548})
      ({\xpos - 0.04 * \xscale}, {\ypos - .101688034})
      ({\xpos - 0.02 * \xscale}, {\ypos - .077218165})
      ({\xpos + 0 * \xscale}, {\ypos - .030109446})
      ({\xpos + 0.02 * \xscale}, {\ypos + .035699141})
      ({\xpos + 0.04 * \xscale}, {\ypos - .048932152})
      ({\xpos + 0.06 * \xscale}, {\ypos - .187548127})
      ({\xpos + 0.08 * \xscale}, {\ypos - .029374102})
      ({\xpos + 0.1 * \xscale}, {\ypos - .124032291})
      ({\xpos + 0.12 * \xscale}, {\ypos + .009788712})
      ({\xpos + 0.14 * \xscale}, {\ypos + .125589231})
      ({\xpos + 0.16 * \xscale}, {\ypos - .058046713})
      ({\xpos + 0.18 * \xscale}, {\ypos + .224891584})
      ({\xpos + 0.2 * \xscale}, {\ypos + .016389727})
      ({\xpos + 0.22 * \xscale}, {\ypos - .109320842})
      ({\xpos + 0.24 * \xscale}, {\ypos - .06390652})
      ({\xpos + 0.26 * \xscale}, {\ypos + .091489909})
      ({\xpos + 0.28 * \xscale}, {\ypos - .053913385})
      ({\xpos + 0.3 * \xscale}, {\ypos + .014361603})
    };

      \draw[-stealth,gray] (\xpos+\xshiftarrowa,\ypos+\yshiftarrowa) -- (\xpos+\xshiftarrowb,\ypos+\yshiftarrowb);
      \node[font=\scriptsize, text=gray] at (\xpos-0.0+\xshiftnoise,\ypos+0.35+\yshiftnoise) {$\epsilon_\mathrm{int}$};
    }

\end{tikzpicture}%
    }
    \vspace{-14pt}
    \caption[Illustration of heat model]{Illustration of the proposed stochastic graph heat model. The model is shown for a six-node undirected (symmetric) graph. Heat diffuses from each node to its neighbours (red arrows); for clarity, diffusion is illustrated only for node 3. The amount of transferred heat  depends on both the connection strength (line thickness) and the temperature gradient between the nodes. Noise is continuously added to the nodes, preventing the system from reaching thermal equilibrium.}
    \label{fig:heat_model}
\end{figure}
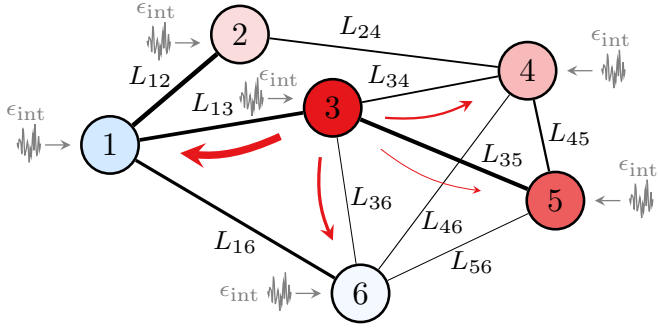

To overcome these limitations, we previously introduced a connectivity estimation method based on stochastic graph heat modelling \cite{goerttler2024stochastic}, which explicitly captures directed and multivariate connectivity structures in time (Figure~\ref{fig:heat_model}).
Specifically, the model assumes a heat diffusion–like spread of activity over the connectivity graph and is driven by stochastic noise. 
While these assumptions are motivated by the connectivity structure in the brain \cite{atasoy2016human} and the widespread occurrence of neuronal noise \cite{guo2018functional}, we do not claim that cortical activity strictly obeys a heat equation. 
Rather, we use stochastic heat diffusion as a tractable phenomenological model of smooth activity propagation on an interaction graph, suitable for connectivity retrieval.
Our approach builds on the graph heat modelling framework of Thanou \textit{et al.} \cite{thanou2017learning}, but relies on stochastic noise instead of source perturbations to infer connectivity. At the same time, it is closely related to linear inverse modelling in oceanography \cite{penland1993prediction}, where systems are similarly modelled as driven by stochastic dynamics. 
However, in contrast to our method, the linear system matrix retrieved in that framework is not interpreted as an estimate of effective connectivity, but rather used to identify principal oscillation patterns.
Our method also bears resemblance to dynamic causal modelling (DCM) frameworks in neurophysiology \cite{friston2003dynamic}, but it is considerably simpler and more computationally tractable \cite{daunizeau2011dynamic}. It is also conceptually related to whole-brain dynamical modelling approaches \cite{deco2011emerging, breakspear2017dynamic}, which simulate large-scale brain activity using coupled neural mass or field models, but, unlike these models, our method directly estimates connectivity from observed signals rather than generating full-scale simulations of whole-brain dynamics.

In this study, we substantially improve the heat-based connectivity measure in two ways. First, we generalise the model by relaxing constraints on both internal and external noise; second, we enhance numerical stability through regularisation of the matrix inverse while preserving the relative graph structure.
We further strengthen the validation of the proposed method using both simulations and real-world datasets. In simulations, we demonstrate the effectiveness of the approach in recovering known connectivity patterns. In particular, we show that the method reliably captures directed connectivity and remains robust to spurious correlations. Our first experiment, conducted on two neurodegenerative disease \gls{EEG} datasets, shows that the proposed measure produces more discriminative connectivity-derived features compared to baseline methods. In a second experiment, based on two simultaneous \gls{EEG}–\gls{fMRI} recordings, we illustrate that the measure shows stronger alignment with a cross-modal reference.

Our contributions in this work can be summarised as follows:
\begin{itemize}
    \item We improve the heat-based connectivity measure by generalising the noise assumptions and enhancing robustness via matrix inverse regularisation.
    \item We characterise the ability of the method to recover directed, multivariate connectivity patterns in simulations and show that the method is robust to spurious coupling.
    \item We demonstrate rapid retrieval of condition-specific connectivity patterns in neurodegenerative disease datasets.
    \item We demonstrate cross-modal connectivity detection using simultaneous \gls{EEG}–\gls{fMRI} recordings.
\end{itemize}

\section{Heat Diffusion Graph Retrieval}
\subsection{Stochastic graph heat modelling-based signal}
A multivariate signal governed by the stochastic graph heat equation with external measurement errors is given as \cite{goerttler2024stochastic}:
\begin{align}
    \mathbf{x}'(t+\Delta t) =& e^{-\Delta t  \mathbf{L}}\left(\mathbf{x}(t) + \boldsymbol{\epsilon}_\mathrm{int}\right) + \boldsymbol{\epsilon}_\mathrm{ext},\label{eq:simulation}
\end{align}
where the internal noise is sampled as $\boldsymbol{\epsilon}_\mathrm{int} \sim \mathcal{N}(\mathbf{0}, \sigma_\mathrm{int}\Delta t)$ and the external noise as $\boldsymbol{\epsilon}_\mathrm{ext} \sim \mathcal{N}(\mathbf{0}, \sigma_\mathrm{ext})$.
The measurement error captures sensor-based noise which is not propagated by the graph connectivity of the system.

The sampled spatial signals can be concatenated to yield the full multivariate signal as a data matrix $\mathbf{X}'\in \mathbb{R}^{N_c \times N_t}$:
\begin{align}
    \mathbf{X}' = \begin{bmatrix}
        \mathbf{x}'(t_0),\mathbf{x}'(t_0 + \Delta t), ..., \mathbf{x}'(t_0 + (N_t - 1)\Delta t)
    \end{bmatrix}.
\end{align}
In the following calculations, we adopt Python-style slicing notation to denote column selection. Generally, $\mathbf{X}_{i{:}j}$ selects columns from index $i$ (inclusive) to $j$ (exclusive), using 0-based indexing and negative indices for wraparound.
For example, $\mathbf{X}_{1{:}}$ denotes all columns of $\mathbf{X}$ except the first one, while $\mathbf{X}_{{:}\shortminus1}$ denotes all columns except the last one.
With these, the vector Equation \eqref{eq:simulation} can be formulated as a matrix equation:
\begin{align}
    \mathbf{X}_{1:}' &= e^{-\Delta t  \mathbf{L}}\left(\mathbf{X}_{:\shortminus1} + \bm{\mathcal{E}}_{:\shortminus1}^\mathrm{int}\right) + \bm{\mathcal{E}}_{1:}^\mathrm{ext} \label{eq:matrix_eq1}\\
    &=e^{-\Delta t  \mathbf{L}}\left(\mathbf{X}_{:\shortminus1}' - \bm{\mathcal{E}}_{:\shortminus1}^\mathrm{ext} + \bm{\mathcal{E}}_{:\shortminus1}^\mathrm{int}\right) + \bm{\mathcal{E}}_{1:}^\mathrm{ext}\label{eq:matrix_eq},
\end{align}
where the noise matrices $\bm{\mathcal{E}}^\mathrm{int}\sim\mathcal{N}(\mathbf{0}_{N_c\times N_t},\sigma_\mathrm{int}\Delta t)$
and $\bm{\mathcal{E}}^\mathrm{ext}\sim\mathcal{N}(\mathbf{0}_{N_c\times N_t},\sigma_\mathrm{ext})$ are sampled from multivariate normal distributions. Lastly, we define the square $(\cdot)^2$ of a matrix $\mathbf{X}$ as $\mathbf{X}^2\coloneqq\mathbf{X}\mathbf{X}^\top$.


\subsection{Computation of Laplacian Matrix}
\label{sec:heat_retrieval}

In this section, we retrieve an algebraic heat graph from a multivariate signal governed by heat-like graph dynamics described in section \ref{ssec:stochastic_graph_heat}.
The principal goal is to retrieve the graph Laplacian $\mathbf{L}$ by inverting Equation \eqref{eq:matrix_eq}.
While only the recorded signal $\mathbf{X}'$ is given, assumptions about $\sigma_\mathrm{int}\Delta t$ and $\sigma_\mathrm{ext}$ as well as noise term averages allow to infer the graph structure and the scale of the heat processing.

To simplify our calculations, we define:
\begin{align}
    \mathbf{M}_{1} &\coloneqq \mathbf{X}_{1:}'-\bm{\mathcal{E}}_{1:}^\mathrm{ext}\\
    \mathbf{M}_{0} &\coloneqq \mathbf{X}_{:\shortminus1}' - \bm{\mathcal{E}}_{:\shortminus1}^\mathrm{ext} + \bm{\mathcal{E}}_{:\shortminus1}^\mathrm{int},
\end{align}
We then rewrite Equation \eqref{eq:matrix_eq} with these two definitions and solve the equation for $\mathbf{L}$, yielding:
\begin{align}
    &&\mathbf{M}_1 &= e^{-\Delta t \mathbf{L}}\mathbf{M}_0&&&\\
    \Rightarrow && \mathbf{M}_1 \mathbf{M}_0^\top \left(\mathbf{M}_0 \mathbf{M}_0^\top\right)^{-1} &= e^{-\Delta t \mathbf{L}}&&&\\
    \Rightarrow && \mathbf{L} = -\frac{1}{\Delta t}\log \bigg(\Big(&\mathbf{M}_1 \mathbf{M}_0^\top\Big) \left({\mathbf{M}_0{\phantom{'}}}^{\hspace{-6pt}2}\right)^{-1} \bigg).\label{eq:L_formula}&&&
\end{align}

This leaves the computation of $\mathbf{M}_1 \mathbf{M}_0^\top$ and $\mathbf{M}_0 \mathbf{M}_0^\top$. We can neglect products of matrices $\mathbf{A}$ with uncorrelated zero-mean noise matrices $\bm{\mathcal{E}}$ given $\mathbb{E}\left[\bm{\mathcal{E}}\mathbf{A}^\top\right] = \bm{0}$, and approximate $\mathbf{X}_{1:}'\bm{\mathcal{E}}_{:\shortminus1}^{\mathrm{int}\top}\approx {\bm{\mathcal{E}}_{:\shortminus1}^\mathrm{int}}^2$ following Equation~\eqref{eq:matrix_eq1}, yielding the following two approximations:
\begin{align}
    \mathbf{M}_1 \mathbf{M}_0^\top =& \mathbf{X}_{1:}'\mathbf{X}_{:\shortminus1}'^\top- \mathbf{X}_{1:}'\bm{\mathcal{E}}_{:\shortminus1}^{\mathrm{ext}\top}+\mathbf{X}_{1:}'\bm{\mathcal{E}}_{:\shortminus1}^{\mathrm{int}\top}\nonumber\\&-\bm{\mathcal{E}}_{1:}^\mathrm{ext}\mathbf{X}_{:\shortminus1}'^\top+\bm{\mathcal{E}}_{1:}^\mathrm{ext}\bm{\mathcal{E}}_{:\shortminus1}^{\mathrm{ext}\top}-\bm{\mathcal{E}}_{1:}^\mathrm{ext}\bm{\mathcal{E}}_{:\shortminus1}^{\mathrm{int}\top}\\\approx& \mathbf{X}_{1:}'\mathbf{X}_{:\shortminus1}'^\top + {\bm{\mathcal{E}}_{:\shortminus1}^\mathrm{int}}^2,\label{eq:M1M0}\\
    {\mathbf{M}_0{\phantom{'}}}^{\hspace{-6pt}2} \approx& {\mathbf{X}_{:\shortminus1}'}^{\hspace{-7pt}2} + {\bm{\mathcal{E}}_{:\shortminus1}^\mathrm{ext}}^2 + {\bm{\mathcal{E}}_{:\shortminus1}^\mathrm{int}}^2.\label{eq:M02}
\end{align}

Lastly, the contribution of individual column pairs to the matrix square can be neglected for increasing numbers of time samples, such that we can approximate:
\begin{align}                        {\bm{\mathcal{E}}_{:}^2}\,\approx{\bm{\mathcal{E}}_{:\shortminus1}^{\phantom{0}}}^{\hspace{-10pt}2}\,\approx{\bm{\mathcal{E}}_{1:}^{\phantom{0}}}^{\hspace{-5pt}2}\approx{\bm{\mathcal{E}}_{:\shortminus2}^{\phantom{0}}}^{\hspace{-10pt}2}\hspace{0.2em}
\approx{\bm{\mathcal{E}}_{2:}^{\phantom{0}}}^{\hspace{-5pt}2}.
\end{align}

%

\subsection{Equal variance assumption}
The noise terms ${\bm{\mathcal{E}}_{:\shortminus1}^\mathrm{int}}^2$ and ${\bm{\mathcal{E}}_{:\shortminus1}^\mathrm{ext}}^2$ in Equations \eqref{eq:M1M0} and \eqref{eq:M02} are not explicitly known and have to be approximated. In our previous work, we used the assumption that $\sigma_\mathrm{int}\approx \sigma_\mathrm{ext}$, or ${\bm{\mathcal{E}}_{:}^\mathrm{ext}}^2 \approx {\bm{\mathcal{E}}_{:}^\mathrm{int}}^2\approx {\bm{\mathcal{E}}_{:}^\mathrm{uni}}^2$, which allowed us to leverage the squared difference to estimate the noise terms:
\begin{align}
    \left(\mathbf{X}_{1:}' - \mathbf{X}_{:\shortminus1}'\right)^2 &\approx {\bm{\mathcal{E}}_{:\shortminus1}^\mathrm{ext}}^2 + {\bm{\mathcal{E}}_{:\shortminus1}^\mathrm{int}}^2 + {\bm{\mathcal{E}}_{1:}^\mathrm{ext}}^2\\ &\approx 2{\bm{\mathcal{E}}_{:}^\mathrm{ext}}^2 + {\bm{\mathcal{E}}_{:}^\mathrm{int}}^2 \approx 3{\bm{\mathcal{E}}_{:}^\mathrm{uni}}^2. &&&\label{eq:1storder}
\end{align}
Simple algebraic manipulations yields the equal variance-based Laplacian solely in terms of the recorded data matrix $\mathbf{X}'$ and the sampling resolution $\Delta t$:
\begin{align}
        \mathbf{L}_\mathrm{EV} \approx -\frac{1}{\Delta t}\log \biggl(&\Bigl(\mathbf{X}_{1:}' \mathbf{X}_{:\shortminus1}'^\top + \frac{1}{3}\left(\mathbf{X}_{1:}' - \mathbf{X}_{:\shortminus1}'\right)^{2}\Bigr) \nonumber\\ &\Bigl({\mathbf{X}_{:\shortminus1}'}^{\hspace{-7pt}2} + \frac{2}{3}\left(\mathbf{X}_{1:}' - \mathbf{X}_{:\shortminus1}'\right)^{2}\Bigr)^{-1} \biggr)\label{eq:heat_L_EV}.
\end{align}

\subsection{Second-order estimation}
We here introduce an alternative approximation of the noise terms, which does not rely on assumptions about the ratio between external and internal noise. This advantage, however, comes at the cost of requiring higher sampling rates.
Specifically, the method leverages the second-order difference $\mathbf{X}_{2:}' - \mathbf{X}_{:\shortminus2}'$ to estimate the noise terms in addition to the first-order difference.
Following Equation \eqref{eq:matrix_eq} and assuming sufficiently small $\Delta t$ such that $e^{-\Delta t  \mathbf{L}}\approx \mathbbm{1}$, the matrix $\mathbf{X}_{2:}'$ can be approximated as:
\begin{align}
    \mathbf{X}_{2:}'=&e^{-\Delta t  \mathbf{L}}\left(e^{-\Delta t  \mathbf{L}}\left(\mathbf{X}_{:\shortminus2}' - \bm{\mathcal{E}}_{:\shortminus2}^\mathrm{ext} + \bm{\mathcal{E}}_{:\shortminus2}^\mathrm{int}\right) + \bm{\mathcal{E}}_{1:\shortminus1}^\mathrm{int}\right) + \bm{\mathcal{E}}_{2:}^\mathrm{ext}\\\approx& \mathbf{X}_{:\shortminus2}' - \bm{\mathcal{E}}_{:\shortminus2}^\mathrm{ext} + \bm{\mathcal{E}}_{:\shortminus2}^\mathrm{int} + \bm{\mathcal{E}}_{1:\shortminus1}^\mathrm{int} + \bm{\mathcal{E}}_{2:}^\mathrm{ext}&&&\\
        \Rightarrow \,\, \mathbf{X}&_{2:}' - \mathbf{X}_{:\shortminus2}' \approx - \bm{\mathcal{E}}_{:\shortminus2}^\mathrm{ext} + \bm{\mathcal{E}}_{:\shortminus2}^\mathrm{int} + \bm{\mathcal{E}}_{1:\shortminus1}^\mathrm{int} + \bm{\mathcal{E}}_{2:}^\mathrm{ext}.&&&
\end{align}
Squaring both sides and removing non-correlated noise products yields:
\begin{align}
    \left(\mathbf{X}_{2:}' - \mathbf{X}_{:\shortminus2}'\right)^2 &\approx {\bm{\mathcal{E}}_{2:}^\mathrm{ext}}^2 + {\bm{\mathcal{E}}_{:\shortminus2}^\mathrm{ext}}^2 + {\bm{\mathcal{E}}_{:\shortminus2}^\mathrm{int}}^2 + {\bm{\mathcal{E}}_{1:\shortminus1}^\mathrm{int}}^{\hspace{-4pt}2} \\
    &\approx 2{\bm{\mathcal{E}}_{:}^\mathrm{ext}}^2 + 2{\bm{\mathcal{E}}_{:}^\mathrm{int}}^2, \label{eq:2ndorder}
\end{align}
which can be used to approximate the noise terms in ${\mathbf{M}_0{\phantom{'}}}^{\hspace{-6pt}2}$ in Equation \eqref{eq:M02} as follows:
\begin{align}
    &\,\,\,\,\,\,\,\,\,\,\frac{1}{2}\left(\mathbf{X}_{2:}' - \mathbf{X}_{:\shortminus2}'\right)^2 \approx {\bm{\mathcal{E}}_{:}^\mathrm{ext}}^2 + {\bm{\mathcal{E}}_{:}^\mathrm{int}}^2 \approx {\bm{\mathcal{E}}_{:\shortminus1}^\mathrm{ext}}^2 + {\bm{\mathcal{E}}_{:\shortminus1}^\mathrm{int}}^2.
\end{align}

To approximate ${\bm{\mathcal{E}}_{:\shortminus1}^\mathrm{int}}^2$ in $\mathbf{M}_1 \mathbf{M}_0^\top$, Equation \eqref{eq:1storder} can be subtracted from Equation \eqref{eq:2ndorder}, yielding:
\begin{align}
    &\left(\mathbf{X}_{2:}' - \mathbf{X}_{:\shortminus2}'\right)^2 - \left(\mathbf{X}_{1:}' - \mathbf{X}_{:\shortminus1}'\right)^2 \approx {\bm{\mathcal{E}}_{:}^\mathrm{int}}^2 \approx {\bm{\mathcal{E}}_{:\shortminus1}^\mathrm{int}}^2.
\end{align}
Finally, these approximations for the two terms $\mathbf{M}_1 \mathbf{M}_0^\top$ and ${\mathbf{M}_0{\phantom{'}}}^{\hspace{-6pt}2}$ can be used to compute the second-order estimation-based Laplacian matrix as follows:
\begin{align}
    \mathbf{L}_2 \approx -\frac{1}{\Delta t}\log \biggl(&\hspace{-0.5pt}\Bigl(\mathbf{X}_{1:}' \mathbf{X}_{:\shortminus1}'^\top \hspace{-1pt}+\hspace{-0.5pt} \left(\mathbf{X}_{2:}' \hspace{-1pt}-\hspace{-0.5pt} \mathbf{X}_{:\shortminus2}'\right)^2 \hspace{-0.75pt}-\hspace{-0.25pt} \left(\mathbf{X}_{1:}' \hspace{-1pt}-\hspace{-0.5pt} \mathbf{X}_{:\shortminus1}'\right)^2\Bigr) \nonumber\\ &\times \Bigl({\mathbf{X}_{:\shortminus1}'}^{\hspace{-7pt}2} \hspace{1pt}+ \frac{1}{2}\left(\mathbf{X}_{2:}' \hspace{-1pt}- \mathbf{X}_{:\shortminus2}'\right)^2 \Bigr)^{-1} \biggr)\label{eq:heat_L_2}.
\end{align}

\subsection{Noise matrix constraints}
\label{ssec:noise_constraints}
The estimated noise matrices ${\bm{\mathcal{E}}_{:\shortminus1}^\mathrm{int}}^2$ and ${\bm{\mathcal{E}}_{:\shortminus1}^\mathrm{ext}}^2 + {\bm{\mathcal{E}}_{:\shortminus1}^\mathrm{int}}^2$ can be constrained by setting off-diagonal entries to zero. Assuming that the noise sources have equal strength across all graph nodes, the noise matrices can be further constrained by replacing each diagonal entry with the mean value across all diagonal entries.

\subsection{Regularisation of matrix inversion}
\label{ssec:regularisation}
The computation of the Laplacian matrix relies on the computation of the matrix inverse of $\left(\mathbf{X}_{:\shortminus1}' - \bm{\mathcal{E}}_{:\shortminus1}^\mathrm{ext} + \bm{\mathcal{E}}_{:\shortminus1}^\mathrm{int}\right)^2={\mathbf{M}_0{\phantom{'}}}^{\hspace{-6pt}2}$. 
However, this inverse can be highly unstable when the condition number of ${\mathbf{M}_0{\phantom{'}}}^{\hspace{-6pt}2}$ is large, which can be computed using the maximal and minimal singular values $\sigma({\mathbf{M}_0{\phantom{'}}}^{\hspace{-6pt}2})$:
\begin{align}
    \kappa({\mathbf{M}_0{\phantom{'}}}^{\hspace{-6pt}2}) = \frac{\sigma_{\max}({\mathbf{M}_0{\phantom{'}}}^{\hspace{-6pt}2})}{\sigma_{\min}({\mathbf{M}_0{\phantom{'}}}^{\hspace{-6pt}2})}.
\end{align} 

We regularise the matrix inverse of ${\mathbf{M}_0{\phantom{'}}}^{\hspace{-6pt}2}$ by adding a diagonal matrix $\gamma \mathbbm{1}$ before computing the inverse, such that the regularised computation of the Laplacian matrix takes the form:
\begin{align}
    \mathbf{L}_\mathrm{R} &= -\frac{1}{\Delta t}\log \left(\left(\mathbf{M}_1 \mathbf{M}_0^\top\right) \left({\mathbf{M}_0{\phantom{'}}}^{\hspace{-6pt}2}+\gamma \mathbbm{1}\right)^{-1} \right).
\end{align}
Specifically, we set
\begin{align}
    \gamma = \frac{- \sqrt{\kappa\left({\mathbf{M}_0{\phantom{'}}}^{\hspace{-6pt}2}\right)} \sigma_{\min}\left({\mathbf{M}_0{\phantom{'}}}^{\hspace{-6pt}2}\right) +  \sigma_{\max}\left({\mathbf{M}_0{\phantom{'}}}^{\hspace{-6pt}2}\right) }{ \sqrt{\kappa\left({\mathbf{M}_0{\phantom{'}}}^{\hspace{-6pt}2}\right)} - 1},
\end{align}
which means that the updated condition number is simply the square root of the original condition number (see Section \ref{sm:regularisation} in the Supplemental Material).

The regularisation leaves the relative graph structure intact and can be applied to Laplacian matrices computed either under an equal-variance assumption or via second-order estimation methods. 
\section{Methods}
\subsection{Baseline connectivity measures}
\label{ssec:baseline_connectivity}
To evaluate the effectiveness of our proposed functional connectivity estimator, we compare it against ten baseline connectivity measures. These include eight functional connectivity measures, one effective connectivity measure, and one structural connectivity measure.
In the following, $x_i(t)$ and $x_j(t)$ are the respective $i$-th and $j$-th row of the measured multivariate signal $\mathbf{X}$, each representing a time series of length $N$. Furthermore, the phase $\phi_i^{(B)}(t)$ of the $i$-th signal $x_i^{(B)}$, filtered in band $B$, is obtained as $\phi_i^{(B)}(t)=\arctan(\hat{x}_i^{(B)}(t)/x_i^{(B)}(t))$, where $\hat{x}_i^{(B)}(t)$ is the Hilbert-transformed signal. Lastly, $\tilde{x}_i(f)$ denotes the Fourier transformed signal of $x_i(t)$. 
Unless otherwise specified, all connectivity estimators were computed using the \texttt{dyconnmap} package \cite{marimpis2021dyconnmap}.

\textbf{Pearson correlation (Corr)} measures the linear correlation between two signals $x_i$ and $x_j$ and captures the degree to which one signal can be expressed as a linear function of another. It is computed as
\begin{align}
    A^\mathrm{Corr}_{ij} = \frac{\sum_t (x_i(t) - \bar{x_i})(x_j(t) - \bar{x_j})}{\sqrt{\sum_t (x_i(t) - \bar{x_i})^2 \sum_t (x_j(t) - \bar{x_j})^2}} \eqqcolon \mathrm{Corr}(x_i, x_j).
\end{align}
We used a custom vectorised Python implementation to efficiently compute the full adjacency matrix.

The \textbf{\gls{PLI}} \cite{stam2007phase} measures phase synchronisation between signals:
\begin{align}
    A^\mathrm{PLI}_{ij} = \left| \frac{1}{N} \sum_{t=1}^N \text{sgn}(\phi_i^{(\alpha)}(t) - \phi_j^{(\alpha)}(t)) \right|,
\end{align}
where $\phi_i^{(\alpha)}$ is the alpha band phase and $\text{sgn}(\cdot)$ is the sign function. 

The \textbf{\gls{wPLI}} \cite{vinck2011improved}
is a modification of the PLI aimed at reducing the influence of volume conduction:
\begin{align}
    A^\mathrm{wPLI}_{ij} = \frac{1}{N} \sum_{t=1}^N \frac{\text{sgn}(\phi_i^{(\alpha)}(t) - \phi_j^{(\alpha)}(t))}{| \phi_i^{(\alpha)}(t) - \phi_j^{(\alpha)}(t) |}.
\end{align}

\textbf{\Gls{PLV}} quantifies the consistency of phase difference between two signals:
\begin{align}
    A_{ij}^\mathrm{PLV} = \left| \frac{1}{N} \sum_{t=1}^N e^{i(\phi_i(t) - \phi_j(t))} \right|.
\end{align}

\textbf{\Gls{Coh}}
measures the linear relationship between two signals in the frequency domain \cite{nolte2004identifying} and is defined as:
\begin{align}
    A_{ij}^\mathrm{Coh} = \frac{|\langle \tilde{x}_i(f) \cdot \tilde{x}_j^*(f) \rangle_\alpha|^2}{\langle |\tilde{x}_i(f)|^2 \rangle_\alpha \langle |\tilde{x}_j(f)|^2 \rangle_\alpha},
\end{align}
where $*$ denotes the complex conjugation and the angle brackets $\langle \cdot \rangle_\alpha$ represent averaging over frequencies in the $\alpha$ band. 

The \textbf{\gls{iCoh}} is a modification of the coherence and emphasises non-zero-lag synchronisation \cite{nolte2004identifying}:
\begin{align}
    A_{ij}^\mathrm{iCoh} = \frac{\mathrm{Im} \left( \langle \tilde{x}_i(f) \cdot \tilde{x}_j^*(f) \rangle_\alpha \right)}{\sqrt{\langle |\tilde{x}_i(f)|^2 \rangle_\alpha \langle |\tilde{x}_j(f)|^2 \rangle_\alpha}},
\end{align}
where $\mathrm{Im}(\cdot)$ extracts the imaginary component. 

\textbf{\Gls{AEC}} measures the correlation between the amplitude envelopes of two signals:
\begin{align}
    A_{ij}^\mathrm{AEC} = \text{Corr}(\text{env}(x_i(t)), \text{env}(x_j(t))),
\end{align}
where $\text{env}(x(t))=\sqrt{x(t)^2+\hat{x}(t)^2}$ denotes the envelope of the analytic signal $x(t)+i\hat{x}(t)$. 

\textbf{\Gls{MI}} quantifies the amount of information shared between two signals and can capture nonlinear dependencies. 
Following \cite{kraskov2004estimating}, \Gls{MI} is estimated as: 
\begin{align}
    A_{ij}^{\mathrm{MI}(k)} &=\\ \psi&(k) - \frac{1}{N} \sum_{t=1}^N \big[\psi(n_i(t) + 1) + \psi(n_j(t) + 1)\big] + \psi(N)\nonumber.
\end{align}
Here, $n_i(t)$ is the number of neighbours of sample t within distance $\epsilon(t)$ in the $x_i$-space, where $\epsilon(t)$ is the distance to the $k$-th nearest neighbor of the $t$-th sample in the joint space $(x_i, x_j)$, while $\psi(\cdot)$ refers to the digamma function.
We used the \texttt{NPEET} toolbox
to compute the \gls{MI} \cite{ver2000non}.

\textbf{Granger causality} assesses the extent to which past values of source signals $x_j(t)$ can predict the future values of the target signal $x_i(t)$. 
It is both the only multivariate and the only directed effective connectivity estimator evaluated in this work.
We used the \texttt{statsmodel} Python package \cite{seabold2010statsmodels} to compute the Granger causality. We set the maximum lag to 4, except for Dataset~I, where it was set to 2 due to the higher number of channels.

The \textbf{geometric distance} is the only baseline included here which measures the structural connectivity. Unlike the other measures, it is not based on the multivariate signal data, but is instead derived from the sensor locations $\mathbf{r}_i$ in the \gls{EEG} recording setup: 
\begin{align}
    A_{ij}^{\mathrm{geom}(\kappa)} &= \exp\left(-\kappa\frac{\|\mathbf{r}_i - \mathbf{r}_j\|_2}{\bar{d}}\right),\\
    \bar{d} &=\langle \|\mathbf{r}_i - \mathbf{r}_j\|_2\rangle_{i\neq j},
\end{align}
where the double vertical bars $\|\cdot\|_2$ denote the Euclidean distance.
The quantity $\bar{d}$ is the average distance across all sensor pairs and normalises the distances, while the parameter $\kappa$ defines the overall scale.

\subsection{Functional magnetic resonance imaging (fMRI) connectivity}
In this subsection, we retrieve the \gls{fMRI} connectivity for \glspl{ROI} corresponding to the \gls{EEG} sensors. This requires the registration of the \gls{fMRI} scans to a common template, within which the \gls{EEG} \glspl{ROI} are defined.

We define the \gls{EEG} \glspl{ROI} by determining the real-space locations of the sensors, which are positioned on the scalp surface surrounding the brain. To capture \gls{fMRI} activity originating in gray matter, we shift the coordinates inward by a factor of 0.15 along the vector pointing toward the center of the brain, which we define as the midway point between the outermost sensors in each dimension. Subsequently, we determine the voxel space representation of the shifted locations and build spherical neighbourhoods of radius 5 voxels around each location, resulting in 515 voxels for each neighbourhood. 

\begin{figure*}[tbp]
    \centering
    \resizebox{0.7\textwidth}{!}{
    \input{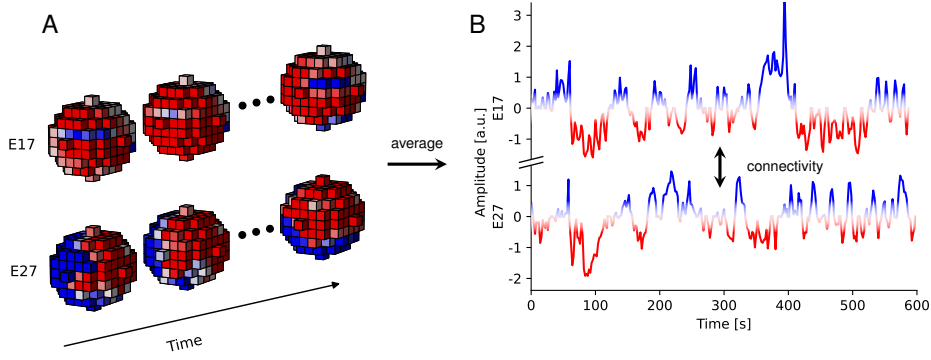}
}
\vspace{-20pt}
    \caption[\Gls{fMRI} connectivity retrieval]{\Gls{fMRI} connectivity estimation for a single EEG sensor pair. (A) Spherical neighbourhoods in the \gls{fMRI} volume, centred on the cortical projections of each EEG sensor, consist of voxels with time-varying signals. The signals within each neighbourhood are spatially averaged to yield (B) a time series for each sensor. Pairwise connectivity is then computed between sensors using Pearson correlation.}
    \label{fig:fmri_connectivity}
\end{figure*}

As illustrated in Figure~\ref{fig:fmri_connectivity}, the \gls{fMRI} values of the \gls{ROI} of each sensor are then averaged across space for each time step, resulting in a multivariate signal composed of time series for each sensor. In the last step, we compute the connectivity between the \glspl{ROI} using Pearson correlation (see Section \ref{ssec:baseline_connectivity}) due to its robustness.

\subsection{Heat model-based connectivity measures}
\label{ssec:heat_conn_measures}

%
\textbf{Heat-EV}, introduced in our earlier work \cite{goerttler2024stochastic}, assumes equal variance between internal and external noise sources and is obtained by computing $\mathbf{L}_\mathrm{EV}$ from Equation \eqref{eq:heat_L_EV}, using the constraints on the noise matrices as described in Section \ref{ssec:noise_constraints}.
Subsequently, the graph Laplacian constraints are enforced \cite{goerttler2024stochastic}, and the adjacency matrix is then extracted from the resulting Laplacian matrix. 

\textbf{Heat-2} is based on second-order noise estimation and differs from Heat-EV in that it uses Equation \eqref{eq:heat_L_2} to compute the Laplacian matrix $\mathbf{L}_\mathrm{2}$. 
We impose the noise matrix constraints followed by the graph Laplacian constraints, yielding the final Laplacian from which the adjacency matrix is derived.

\textbf{Heat-2R} extends the previous Heat-2 measure by additionally incorporating regularisation.
To this end, the computation of $\mathbf{L}_\mathrm{2}$ from Equation \eqref{eq:heat_L_2} is modified by adding the regularisation term $\gamma \mathbbm{1}$ to the argument of the matrix inverse, as specified in Section \ref{ssec:regularisation}. This regularisation is expected to enhance the robustness of the Heat-2 measure in capturing fine-grained structural details.

\subsubsection*{Operating Regime}
\label{ssec:assumptions_regime}
Heat-EV relies on an equal-variance noise assumption, which restricts its use to settings where internal and external noise sources are roughly comparable. Heat-2 removes this restriction by using a second-order noise estimation under a small-$\Delta t$ approximation, but requires a higher sampling rate. 
Unlike Heat-2R, both Heat-EV and Heat-2 preserve the connectivity scale under their respective assumptions, making them suitable for scale estimation.
Heat-2R introduces regularisation in the matrix inversion step, which makes it more applicable to noisy datasets. 

\subsection{Evaluation metrics}
We use three evaluation metrics for the retrieved connectivity measures in our simulations and experiments.

The first evaluation metric is the degree of alignment between an estimated and a reference connectivity measure.
The metric measures the Pearson correlation between the off-diagonal entries of the estimated connectivity measure and the reference connectivity. 
In our simulations, the reference connectivity is the ground truth connectivity matrix used to simulate the data. In Experiment~II, on the other hand, we use the \gls{fMRI} connectivity matrix as the reference connectivity.

The second metric measures the \gls{SVM} classification accuracy of features derived from the upper triangular part of the connectivity matrix. Given the number of true positives $TP_i$ and true negatives $TN_i$ for class $i$, as well as the total number of samples $N_s$, the accuracy is given by $\left(\sum_{i=1}^{K} TP_i + TN_i\right)/N_s$.
%
%
To train the \gls{SVM}, we employed nested cross-validation with an outer stratified 10-fold split to evaluate model performance, and an inner 3-fold stratified cross-validation for hyperparameter tuning. In each outer fold, a pipeline combining PCA and an SVM classifier with radial basis function kernel (RBF) was trained. The number of \gls{PCA} components was tuned over $N_\mathrm{PCA}\in\{N_\mathrm{train}, N_\mathrm{train} - 10, N_\mathrm{train} -20\}$, where $N_\mathrm{train}$ denotes the number of available training samples in the inner loop. The SVM was tuned over regularisation parameters $C\in\{0.1, 1, 10\}$ and kernel bandwidth settings $\gamma \in \{\text{scale}, \text{auto}, 0.01, 0.1\}$, where the settings \textit{scale} and \textit{auto} are based on the \gls{SVM} implementation provided in \texttt{scikit-learn} \cite{pedregosa2011scikit}.
This nested cross-validation procedure was repeated for 100 bootstrap samples, generated via stratified sampling with replacement.
The \gls{SVM}-based classification pipeline was implemented using the \texttt{scikit-learn} library in Python \cite{pedregosa2011scikit}.

Lastly, we evaluate the computation cost of each connectivity measure by measuring the average wall-clock time per sample using the \texttt{perf\_counter()} function from Python's time module (Python 3.11 \cite{python311}). All experiments were conducted on a MacBook Pro featuring an Apple M2 Max chip with 12 CPU cores (8 performance and 4 efficiency cores) and 32 GB of unified memory.

\section{Simulation}
\subsection{Graph structure}
\label{ssec:graph_structure}
In this section, we construct the random graph connectivity matrices that form the basis of our simulations, considering both directed and undirected graphs.

To construct the adjacency matrix of the \textbf{directed graph}, we set the diagonal elements to 1 and sample off-diagonal elements from a uniform distribution between 0 and 0.8:
\begin{align}
A_{ij}^{\mathrm{d}} =
\begin{cases}
1 & \text{if } i = j, \\
\sim\mathcal{U}_{[0, 0.8]} & \text{if } i \neq j.
\end{cases}
\end{align}

We derive an \textbf{undirected graph} $\mathbf{A}^{\mathrm{ud}}$ from the directed graph $\mathbf{A}^{\mathrm{d}}$ by symmetrising it and enforcing positive semi-definiteness (see Section \ref{sec:der_ud} in the Supplemental Material for details).

\subsection{Heat-based simulation}
\label{ssec:heat_simulation}
Equation \eqref{eq:simulation} describes the temporal evolution of a graph signal resulting from heat-like propagation, which enables the sequential simulation of multivariate signals.
The equation contains three parameters: The internal noise scale $\sigma_\mathrm{int}$, the external noise scale $\sigma_\mathrm{ext}$, and the graph thermal diffusivity $\alpha$.
The simulation of a subsequent time step requires a base Laplacian matrix $\mathbf{L}_0$ and an initial graph signal $\mathbf{x}_0$ at $t=t_0$. 
The Laplacian can be derived from both a directed or undirected adjacency matrix $\mathbf{A}$.
In addtion, $\mathbf{x}_0$ can be initialised by sampling from a standard normal distribution. 

For simulations with $N_t\gg0$, the simulation can suffer from positive feedback loops. 
We therefore incorporate a saturation mechanism into Equation \eqref{eq:simulation} to prevent large signal values:
\begin{align}
    \mathbf{x}'(t+\Delta t) =& a\,\mathrm{tanh}\left(\frac{e^{-\Delta t\hspace{0.05em} \alpha  \mathbf{L}_0}\left(\mathbf{x}(t) + \boldsymbol{\epsilon}_\mathrm{int}\right) + \boldsymbol{\epsilon}_\mathrm{ext}}{a}\right),\label{eq:simulation_mod}
\end{align}
where $a$ controls the extent of the saturation, with smaller values corresponding to tighter saturation. In our simulation, we set $a=100$, corresponding to mild saturation control.

In addition, the simulation requires a warm-up phase for the system dynamics to stabilise, given its random initialisation. We therefore excluded the first 100 simulated time samples.


\subsection{Spurious relationship simulation}
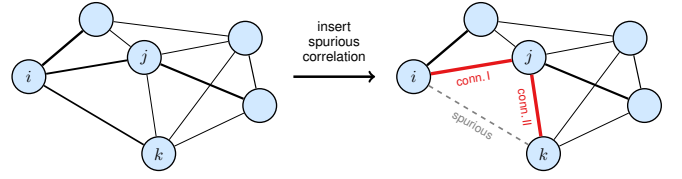
\begin{figure}[tbh]
    \centering
    \vspace{10pt}
    \resizebox{1\columnwidth}{!}{%
    \renewcommand*\familydefault{\sfdefault} 
\fontfamily{\sfdefault}\selectfont

\noindent
\begin{tikzpicture}[bend angle=10]
    \clip (1., -2.2) rectangle (14.7, 1.6);
  \node[draw, circle, fill=dodgerblue!20, thick, minimum size=0.7cm] (t0) at (1.5,0){$i$};
  \node[draw, circle, fill=dodgerblue!20, thick, minimum size=0.7cm,shift={(1.4, 1.19)}] (t1) at (t0){};
  \node[draw, circle, fill=dodgerblue!20, thick, minimum size=0.7cm,shift={(2.7, -1.6)}] (t2) at (t0){$k$};
  \node[draw, circle, fill=dodgerblue!20, thick, minimum size=0.7cm,shift={(2.4, 0.4)}] (t3) at (t0){$j$};
  \node[draw, circle, fill=dodgerblue!20, thick, minimum size=0.7cm,shift={(4.5, 0.8)}] (t4) at (t0){};
  \node[draw, circle, fill=dodgerblue!20, thick, minimum size=0.7cm,shift={(4.8,-0.6)}] (t5) at (t0){};

  \draw[-,thick,line width=0.5mm] (t1) edge (t0) node[text width=0.8cm,align=center, font={\tiny}, midway,below] {};
  \draw[-,thick,line width=0.35mm] (t0) -- (t2) node[text width=0.8cm,align=center, font={\tiny}, midway,below] {};
  \draw[-,thick,line width=0.4mm] (t3) -- (t0) node[text width=0.8cm,align=center, font={\tiny}, midway,below] {};
  \draw[-,thick,line width=0.2mm] (t3) -- (t4) node[text width=0.8cm,align=center, font={\tiny}, midway,below] {};
  \draw[-,thick,line width=0.15mm] (t3) edge (t1) node[text width=0.8cm,align=center, font={\tiny}, midway,below] {};
  \draw[-,thick,line width=0.2mm] (t1) edge (t4) node[text width=0.8cm,align=center, font={\tiny}, midway,below] {};
  \draw[-,thick,line width=0.08mm] (t2) edge (t3) node[text width=0.8cm,align=center, font={\tiny}, midway,below] {};
  \draw[-,thick,line width=0.45mm] (t5) edge (t3) node[text width=0.8cm,align=center, font={\tiny}, midway,below] {};
  \draw[-,thick,line width=0.10mm] (t2) edge (t5) node[text width=0.8cm,align=center, font={\tiny}, midway,below] {};
  \draw[-,thick,line width=0.25mm] (t4) edge (t5) node[text width=0.8cm,align=center, font={\tiny}, midway,below] {};
  \draw[-,thick,line width=0.12mm] (t2) edge (t4) node[text width=0.8cm,align=center, font={\tiny}, midway,below] {};

  \draw[->, thick, line width=1.5pt] (7,0) -- (8.7,0) node[midway, above=5pt,text width=2.1cm,align=center] {\footnotesize insert\\spurious\\[-3pt]correlation};
  
  \node[draw, circle, fill=dodgerblue!20, thick, minimum size=0.7cm] (t0) at (9.5,0){$i$};
  \node[draw, circle, fill=dodgerblue!20, thick, minimum size=0.7cm,shift={(1.4, 1.19)}] (t1) at (t0){};
  \node[draw, circle, fill=dodgerblue!20, thick, minimum size=0.7cm,shift={(2.7, -1.6)}] (t2) at (t0){$k$};
  \node[draw, circle, fill=dodgerblue!20, thick, minimum size=0.7cm,shift={(2.4, 0.4)}] (t3) at (t0){$j$};
  \node[draw, circle, fill=dodgerblue!20, thick, minimum size=0.7cm,shift={(4.5, 0.8)}] (t4) at (t0){};
  \node[draw, circle, fill=dodgerblue!20, thick, minimum size=0.7cm,shift={(4.8,-0.6)}] (t5) at (t0){};

  \draw[-,thick,line width=0.5mm] (t1) edge (t0) node[text width=0.8cm,align=center, font={\tiny}, midway,below] {};
  \draw[-,thick,line width=0.35mm,dashed,draw=gray] (t0) -- (t2) node[text width=1.5cm,align=center,sloped,gray, font={\scriptsize}, midway,below] {spurious};
  \draw[-,thick,line width=0.7mm,draw=newred] (t3) -- (t0) node[text width=2cm,align=center,sloped,newred,font={\scriptsize}, midway,below] {conn.\,I};
  \draw[-,thick,line width=0.7mm,draw=newred] (t3) -- (t2) node[text width=2cm,align=center,sloped,newred,font={\scriptsize}, midway,below] {conn.\,II};
  \draw[-,thick,line width=0.2mm] (t3) -- (t4) node[text width=0.8cm,align=center, font={\tiny}, midway,below] {};
  \draw[-,thick,line width=0.15mm] (t3) edge (t1) node[text width=0.8cm,align=center, font={\tiny}, midway,below] {};
  \draw[-,thick,line width=0.2mm] (t1) edge (t4) node[text width=0.8cm,align=center, font={\tiny}, midway,below] {};
  \draw[-,thick,line width=0.45mm] (t5) edge (t3) node[text width=0.8cm,align=center, font={\tiny}, midway,below] {};
  \draw[-,thick,line width=0.10mm] (t2) edge (t5) node[text width=0.8cm,align=center, font={\tiny}, midway,below] {};
  \draw[-,thick,line width=0.25mm] (t4) edge (t5) node[text width=0.8cm,align=center, font={\tiny}, midway,below] {};
  \draw[-,thick,line width=0.12mm] (t2) edge (t4) node[text width=0.8cm,align=center, font={\tiny}, midway,below] {};

\end{tikzpicture}%
    }
    \vspace{-10pt}
    \caption[Insertion of spurious correlation]{Insertion of spurious correlation between two arbitrary nodes $i$ and $k$. The direct connection between $i$ and $k$ is removed, while the connection between $i$ and $k$ through an intermediate node $j$ is increased. All other connections remain the same.}
    \label{fig:spurious_theory}
\end{figure}

We now simulate a multivariate signal exhibiting a spurious relationship between two nodes, i.e., a statistical association induced by a third node rather than a direct interaction. The signal is generated from the heat simulation as described in Section \ref{ssec:heat_simulation}, using the adjacency matrix $\mathbf{A}^\mathrm{ud}$ of a randomly generated undirected graph. 
Figure~\ref{fig:spurious_theory} illustrates how the spurious relationship is inserted into the undirected graph. 
To this end, three nodes are picked at random, namely a source node $i$, an intermediate node $j$, and a target node $k$. The connection between the source and the target node is set to 0 ($A_{ik}^\mathrm{ud}=A_{ki}^\mathrm{ud}=0$), while the two connections via the intermediate node are set to a value close to 1 ($A_{ij}^\mathrm{ud}=A_{ji}^\mathrm{ud}=0.8$ and $A_{jk}^\mathrm{ud}=A_{kj}^\mathrm{ud}=0.8$).
The signal is then generated with $\alpha=0.3$, $\sigma_\mathrm{int}=1$, and $\sigma_\mathrm{ext}=1$, corresponding to the standard simulation parameters.

\subsection{Simulation procedure}
\label{ssec:sim_procedure}

We simulated multivariate signals based on the heat model with $N = 32$ nodes over $N_t = 15,000$ time points, corresponding to a typical \gls{EEG} recording of 5 minutes at a sampling rate of 50\,Hz.
%
For both the undirected-graph and directed-graph simulations, we considered 25 parameter configurations by generating five logarithmically spaced values for $\alpha \in [0.01, 0.1]$ and $\sigma^{\mathrm{int}} \in [0.1, 10]$,  while keeping $\sigma^{\mathrm{ext}} = 1$ fixed. 
Furthermore, for the artificial correlation-based simulation, we varied the maximum frequency values as $f_\mathrm{max} = 1,2,\dots,25$.

We simulated each configuration $N_\mathrm{rep}=100$ times. We evaluated the connectivity measures in the heat-based and correlation-based simulations by examining their mean alignment with the ground truth adjacency matrix $\mathbf{A}$ used in the simulation. 
To reduce computational cost, we evaluated Granger causality and \gls{MI} using only the first 10 simulations for the heat-based simulation.
To evaluate the ability of the connectivity measures to detect the simulated spurious relationships, we computed the mean estimated value of the spurious connection and of the two connections via the intermediate node.
%

\section{Datasets}
\subsection{Datasets I-II: neurodegenerative disorders (EEG)}
\textbf{Dataset~I:} This dataset was collected by Singh \textit{et al.} \cite{singh2023evoked} from 149 individuals, comprising 100 patients with \gls{PD} and 49 \gls{HC}\footnote{\href{https://openneuro.org/datasets/ds004584/versions/1.0.0} {https://openneuro.org/datasets/ds004584/versions/1.0.0}}. Recordings were acquired using a 64-channel BrainVision cap during a two-minute resting-state session, in which participants remained seated in a quiet room with their eyes open. All \gls{EEG} channels were referenced to the Pz electrode, resulting in 63 channels available for analysis.

\textbf{Dataset~II:} Recorded by Miltiadous \textit{et al.} \cite{miltiadous2023dataset}, this dataset includes 36 \gls{AD} patients, 23 \gls{FTD} patients, and 29 \glspl{HC}\footnote{\href{https://openneuro.org/datasets/ds004504/versions/1.0.6} {https://openneuro.org/datasets/ds004504/versions/1.0.6}}.
It was acquired using an 10-20 \gls{EEG} system with 19 electrodes and two reference electrodes, sampled at a rate of 500$\,$Hz.
Participants were directed to close their eyes throughout the recordings, which lasted for approximately 13-14 minutes on average. 
\gls{MMSE} scores were assessed for each \gls{AD} patient.
%

\subsection{Datasets III-IV: simultaneous EEG-fMRI}
\textbf{Dataset~III:} This publicly available\footnote{\href{https://data.mendeley.com/datasets/crhybxpdy6/2} {https://data.mendeley.com/datasets/crhybxpdy6/2}} simultaneous EEG–fMRI dataset was collected by Gallego-Rudolf \textit{et al.} on 20 healthy male subjects \cite{gallego2023simultaneous}. \Gls{EEG} was recorded using a 32-channel Geodesic Sensor Net and corrected for gradient artifacts, while \gls{fMRI} data were acquired on a 3T GE MR750 scanner with a \gls{TR} of 2\,s. Structural T1-weighted images were also acquired for anatomical reference. Only the simultaneously recorded resting-state data were included in this study.


\textbf{Dataset~IV:} The second simultaneous EEG-fMRI dataset\footnote{\href{https://fcon\_1000.projects.nitrc.org/indi/retro/nat\_view.html}{https://fcon\_1000.projects.nitrc.org/indi/retro/nat\_view.html}} was collected from 22 healthy individuals \cite{telesford2023open}. We used only the resting-state recordings. \Gls{EEG} data were acquired using a 64-channel Brain Products BrainCapMR system and preprocessed using the pipeline provided by the dataset authors. Resting-state BOLD \gls{fMRI} data were acquired on a Siemens TrioTim 3T scanner (\gls{TR}=2.1\,s). High-resolution T1-weighted structural images were also collected.


\subsection{fMRI preprocessing and registration}
Preprocessing of \gls{fMRI} Datasets~III and IV consisted of slice-timing correction using the \texttt{slicetimer} routine of the FSL \gls{fMRI} analysis package\footnote{\href{https://www.fmrib.ox.ac.uk/fsl/slicetimer/index.html} {https://www.fmrib.ox.ac.uk/fsl/slicetimer/index.html}} \cite{jenkinson2002improved,jenkinson2012fsl}, which accounts for acquisition delays between slices.
The data were then registered to a nonlinear, symmetric T1-weighted template\footnote{\href{https://identifiers.org/neurovault.image:29396} {https://identifiers.org/neurovault.image:29396}} \cite{fonov2011unbiased}. 
Specifically, we first registered the functional scan of each participant with their anatomical scan using a rigid transformation, and second the anatomical scan with the standard template using a symmetric normalisation comprising an affine and deformable transformation. Registration was perfomed using the  \texttt{ANTsPy} Python library \cite{avants2011reproducible,antspy}.
Lastly, we applied temporal filtering after registration using a fourth-order Butterworth filter with a cutoff-frequency of 0.01\,Hz.

\subsection{Mapping \gls{EEG} sensors onto the \gls{fMRI} cortical surface}
\begin{figure}[tbp]
\sffamily
    \centering
    \vspace{5pt}
    \begin{tikzpicture}
    \node[anchor=north west] (image) at (0,0) {\includegraphics[width=0.99\columnwidth]{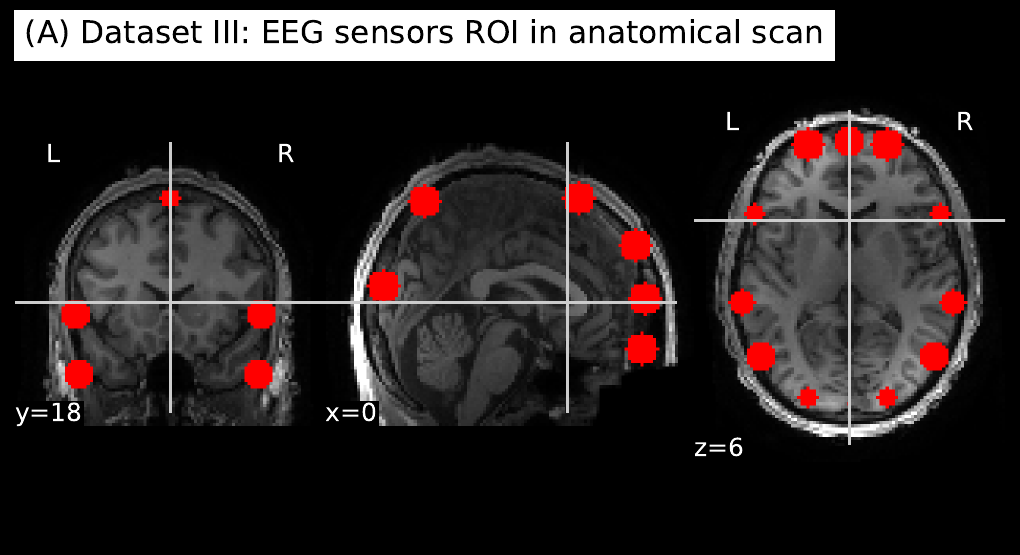}};
    \node[anchor=north west] (image) at (0.004,-4) {\includegraphics[width=0.99\columnwidth, trim={0 0.5cm 0 0},clip]{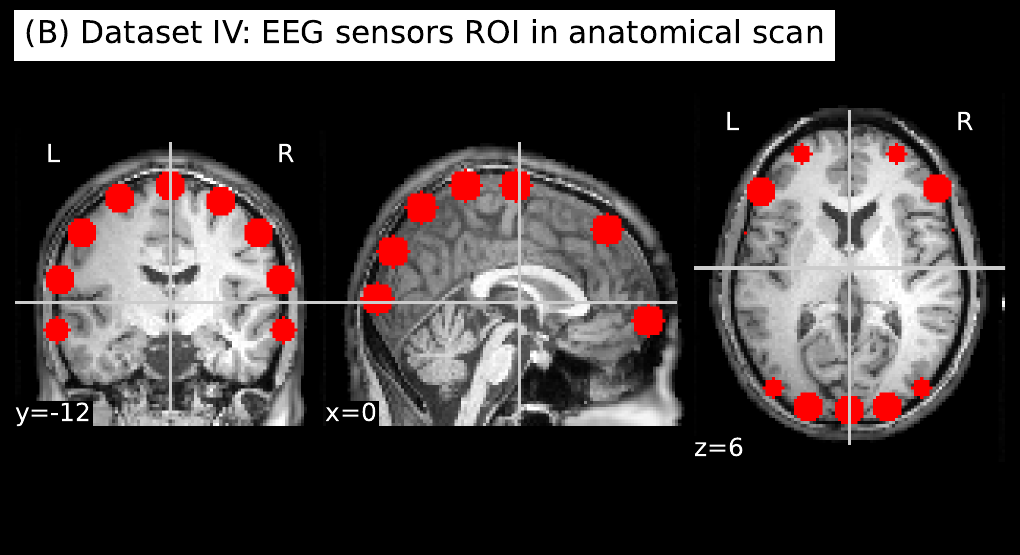}};
        
    \end{tikzpicture}
    \caption[\Glspl{ROI} of \gls{EEG} sensors]{\Glspl{ROI} (in red) of \gls{EEG} sensors for (A) Dataset~III and (B) Dataset~IV. Slice values were chosen to maximise the number of sensors displayed. Note that not all sensors are visible in the two-dimensional slices due to the 3D slicing geometry, and some may appear more than once.}
    \label{fig:roi}
\end{figure}

The \gls{EEG} sensor locations in Datasets~III and IV were registered to the standard \gls{MRI} template and subsequently shifted to corresponding cortex \glspl{ROI}.
We used the Python package \texttt{MNE-python} \cite{gramfort2013meg} to retrieve the sensor locations from the given channels. These locations were already on a compatible scale, so no additional scaling was applied. In the case of Dataset~III, however, the origin of the \gls{EEG} sensor space and that of the \gls{fMRI} standard template where misaligned. We visually aligned the two spaces by shifting the \gls{EEG} sensor space origin by $20$\,mm in $y$-direction and $30$\,mm in $z$-direction. 
Figure~\ref{fig:roi} shows the resulting \glspl{ROI} of the \gls{EEG} sensors for selected brain slices from the registered anatomical scans for both datasets.

\section{Experimental Results}
\label{sec:heat_results}
\subsection{Heat-based simulation}
\begin{figure*}[tbp]
    \centering
    \includegraphics[width=0.9\textwidth]{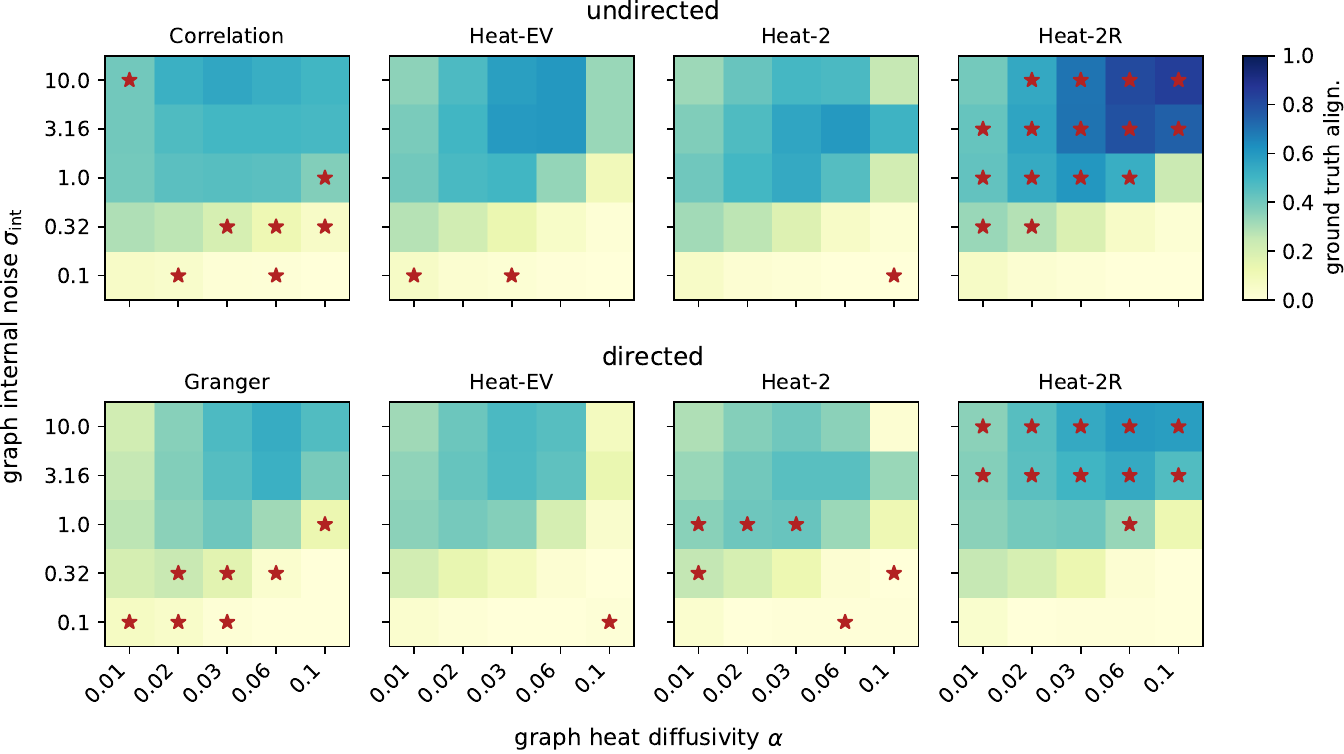}
    \hspace{10pt}
    \caption[Ground truth alignment of the heat-based simulation for four selected connectivity measures]{Mean ground truth alignment of the heat-based simulation for four selected connectivity measures, across all parameter configurations for undirected (first row) and directed (second row) graphs. Red stars indicate the best performance across the four connectivity measures for each graph type. The Heat-2R measure clearly outperforms all other measures, particularly at large values of $\sigma_\mathrm{int}$.}
    \label{fig:sim_heat}
\end{figure*}
Figure~\ref{fig:sim_heat} shows the simulation-averaged ground truth alignment for the best-performing baseline connectivity measure and our three heat connectivity measures described in Section \ref{ssec:heat_conn_measures}. 
For undirected graphs, the correlation measure is the best-performing baseline measure. The regularised heat measure Heat-2R clearly outperforms the correlation measure if the graph heat diffusivity $\alpha$ and the graph internal noise $\sigma_\mathrm{int}$ are sufficiently large. For small values of $\sigma_\mathrm{int}$, the correlation measure outperforms the Heat-2R measure, although the performance of all measures at these values is poor.

In the case of the directed graph, Granger causality is the best-performing baseline, particularly at low graph internal noise values $\sigma_\mathrm{int}$. Similar to the undirected graph, the Heat-2R measure clearly outperforms all other measures at large values of $\sigma_\mathrm{int}$.
Generally, the performance for directed graphs is lower than for undirected graphs.

\subsection{Spurious relationship simulation}
\begin{figure*}[tbp]
    \centering
    \includegraphics[width=0.8\textwidth]{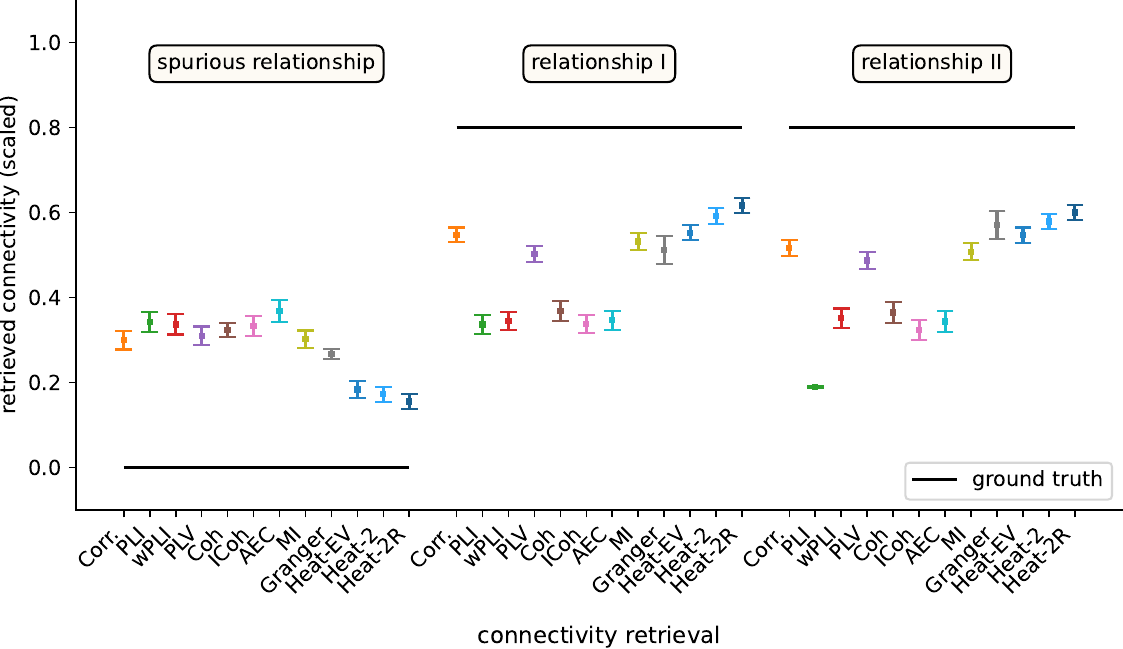}
    \vspace{0pt}
    \caption[Retrieved spurious connectivity values over connectivity measures]{Scaled retrieved connectivity values over all connectivity measures for the spurious connection and the two real intermediate connections. The heat-based connectivity measures Heat-EV (blue), Heat-2 (light blue), and Heat-2R (dark blue) are closest to the ground truth connectivity values of 0 for the spurious and 0.8 for the intermediate connections. The best-performing baselines are Granger causality and correlation.}
    \label{fig:spurious}
\end{figure*}

Figure~\ref{fig:spurious} depicts the mean retrieved connectivity, scaled to match the mean and variance of the ground truth adjacency matrix, for the spurious connection and the two intermediate connections. The results indicate that the Heat-2R measure most closely reproduces the target connection strength of 0 for the spurious connection, as well as the target connection strength of 0.8 for the intermediate connections.
The performance of the Heat-2R measure is closely followed by the two heat-based measures Heat-2 and Heat-EV. 
The heat-based measures outperform all other baseline connectivity measures, with the exception of Granger causality, which achieves a closer match for intermediate connection II than the Heat-EV measure.
The baselines providing the closest match are Granger causality, correlation, \gls{MI} and \gls{PLV}.

\subsection{Experiment~I: classification}

Table \ref{tab:adj_classification} shows the performance of each connectivity measure in terms of accuracy, precision, recall and time duration per sample. 
Apart from accuracy, the Heat-2R connectivity measure also performs well in terms of both precision and recall on both datasets, whereas Heat-EV shows poor precision on Dataset~I.
In terms of computation time, the heat-based measures are also the fastest, followed closely only by Pearson correlation.

\subsection{Experiment~II: fMRI alignment}
In Experiment~II, we retrieved the \gls{EEG} connectivity for the three heat-based measures and ten baseline measures, and computed the alignment with the \gls{fMRI} connectivity.

\begin{table*}[tbp]
    \centering
    \caption[\Gls{SVM} classification performance for Datasets I and II for available connectivity measures]{Experiment I: \Gls{SVM} classification performance of connectivity measure values and connectivity measure computation time. Best-performing values are marked in bold, while second-best performing values are underlined.}
    \vspace{4pt}
    \resizebox{1\textwidth}{!}{\begin{tabular}{llcccScccS}
\toprule
conn. measure & type & \multicolumn{4}{c}{Dataset I: PD \& HC} & \multicolumn{4}{c}{Dataset II: AD, FTD \& HC } \\
\cmidrule(lr){3-6}\cmidrule(lr){7-10} & & acc. [\%]& prec. [\%] & recall [\%] & \multicolumn{1}{c}{time/samp. [s]} & acc. [\%]& prec. [\%] & recall [\%] & \multicolumn{1}{c}{time/samp. [s]} \\
\midrule
Correlation & time-domain & 66.5$\pm$1.0 & 34.4$\pm$1.9 & 49.8$\pm$1.0 & \underline{0.08} & 46.8$\pm$2.5 & 35.6$\pm$3.8 & 41.5$\pm$2.8 & \underline{0.04} \\ 
\cdashlinelr{1-10}
PLI & frequency-domain & 63.8$\pm$2.3 & 41.0$\pm$4.8 & 50.8$\pm$2.3 & 0.35 & 49.4$\pm$2.1 & 35.9$\pm$3.1 & 45.0$\pm$3.4 & 0.31 \\ 
wPLI &  & 66.2$\pm$1.9 & \underline{50.9$\pm$5.7} & \textbf{53.8$\pm$2.9} & 1.15 & 47.1$\pm$3.0 & 36.4$\pm$4.0 & 44.4$\pm$3.9 & 0.52 \\ 
PLV &  & 66.3$\pm$0.8 & 33.9$\pm$1.4 & 49.6$\pm$0.9 & 0.14 & 48.2$\pm$2.9 & 37.2$\pm$4.0 & 43.2$\pm$3.6 & 0.17 \\ 
Coh &  & 65.7$\pm$2.2 & 44.6$\pm$6.0 & 51.8$\pm$2.7 & 2.50 & 52.5$\pm$2.9 & 40.5$\pm$4.3 & 48.8$\pm$4.0 & 1.05 \\ 
ICoh &  & \underline{66.8$\pm$0.7} & 33.9$\pm$1.2 & 50.0$\pm$1.0 & 2.31 & 52.4$\pm$1.8 & 37.2$\pm$2.9 & 48.3$\pm$3.3 & 0.96 \\ 
AEC &  & 66.7$\pm$1.5 & 44.6$\pm$5.1 & 51.8$\pm$1.8 & 0.18 & 52.3$\pm$2.3 & 39.0$\pm$3.4 & 48.0$\pm$3.5 & 0.30 \\ 
\cdashlinelr{1-10}
MI & information-theoretic & \underline{66.8$\pm$1.1} & 36.2$\pm$3.0 & 50.4$\pm$1.3 & 69.90 & 51.8$\pm$3.2 & \textbf{50.0$\pm$5.2} & 49.4$\pm$4.5 & 37.68 \\ 
\cdashlinelr{1-10}
Granger & causal & \underline{66.8}$\pm$0.6 & 33.9$\pm$1.2 & 50.0$\pm$1.0 & 400.26 & 45.2$\pm$2.5 & 33.6$\pm$4.7 & 39.8$\pm$3.1 & 13.60 \\ 
\cdashlinelr{1-10}
\textbf{Heat-EV} & model-based & \underline{66.8$\pm$0.7} & 34.9$\pm$2.4 & 50.1$\pm$1.1 & \textbf{0.05} & \textbf{54.8$\pm$2.9} & 42.7$\pm$3.6 & \textbf{51.8$\pm$3.4} & \underline{0.04} \\ 
\textbf{Heat-2} &  & 66.0$\pm$1.8 & 42.7$\pm$4.8 & 51.5$\pm$2.2 & \textbf{0.05} & 50.9$\pm$2.4 & 42.5$\pm$3.9 & 48.1$\pm$3.6 & \textbf{0.03} \\ 
\textbf{Heat-2R} &  & \textbf{67.9$\pm$1.2} & \textbf{51.7$\pm$5.9} & \underline{53.6$\pm$2.0} & \textbf{0.05} & \underline{53.5$\pm$3.3} & \underline{45.3$\pm$4.1} & \underline{50.4$\pm$3.9} &\textbf{0.03} \\ 
\bottomrule
\end{tabular}
}
    \label{tab:adj_classification}
\end{table*}

Table \ref{tab:fmri_alignment} presents the mean \gls{fMRI} alignment values across both datasets, the computation time per sample, and the $p$-value indicating whether each method's alignment exceeds that of Heat-2R. Heat-2R connectivity achieves mean \gls{fMRI} alignment of 0.20 and 0.31 on Datasets III and IV, respectively, which significantly exceeds all connectivity measures except for coherence and Heat-EV connectivity on Dataset~III, and correlation and geometric connectivity on Dataset~IV. On Dataset~III, only correlation, geometric, and Heat-2 exhibit faster computation times, while on Dataset~IV, only geometric, Heat-EV, and Heat-2 are faster.
\begin{table*}[tbp]
    \centering
    \caption[\Gls{fMRI} alignment of connectivity measures for Datasets III and IV]{Experiment II: \Gls{fMRI} alignment of connectivity measures. Also displayed are the computation time per sample and the $p$-value indicating whether the method significantly exceeds the Heat-2R alignment. Highest alignment is marked in bold, while second-highest alignment is underlined.}
    \vspace{4pt}
    \resizebox{1\textwidth}{!}{
    \begin{tabular}{llcSllcSll}
\toprule
conn. measure & type & \multicolumn{4}{c}{EEG-fMRI Dataset III} & \multicolumn{4}{c}{EEG-fMRI Dataset IV} \\
\cmidrule(lr){3-6}\cmidrule(lr){7-10} & & fMRI align.& \multicolumn{1}{c}{time/samp. [s]} & \multicolumn{2}{c}{$p>$ Heat-2R} & fMRI align. & \multicolumn{1}{c}{time/samp. [s]} & \multicolumn{2}{c}{$p>$ Heat-2R} \\
\midrule
Correlation & time-domain & 0.05 $\pm$ 0.09 & 0.06 & 6.3$\times 10^{9}$ & *** & \textbf{0.32 $\pm$ 0.16} & 0.22 & 0.73 &     \\ 
\cdashlinelr{1-10}
PLI & frequency-domain & 0.02 $\pm$ 0.07 & 0.32 & 1.3$\times 10^{12}$ & *** & $-$0.02 $\pm$ 0.10\phantom{$-$} & 0.82 & $<10^{-12}$\phantom{$<$} & *** \\ 
wPLI &  & $-$0.02 $\pm$ 0.09\phantom{$-$} & 1.03 & $<10^{-12}$\phantom{$<$} & *** & $-$0.02 $\pm$ 0.03\phantom{$-$} & 2.72 & $<10^{-12}$\phantom{$<$} & *** \\ 
PLV &  & 0.03 $\pm$ 0.10 & 0.25 & 4.0$\times 10^{10}$ & *** & $-$0.00 $\pm$ 0.06\phantom{$-$} & 0.45 & $<10^{-12}$\phantom{$<$} & *** \\ 
Coh &  & \underline{0.18 $\pm$ 0.11} & 2.15 & 0.22 &     & 0.04 $\pm$ 0.07 & 5.71 & $<10^{-12}$\phantom{$<$} & *** \\ 
ICoh &  & $-$0.06 $\pm$ 0.11\phantom{$-$} & 2.10 & $<10^{-12}$\phantom{$<$} & *** & $-$0.07 $\pm$ 0.08\phantom{$-$} & 5.57 & $<10^{-12}$\phantom{$<$} & *** \\ 
AEC &  & 0.14 $\pm$ 0.12 & 0.29 & 1.1$\times 10^{2}$ & *   & 0.06 $\pm$ 0.08 & 0.62 & $<10^{-12}$\phantom{$<$} & *** \\ 
\cdashlinelr{1-10}
MI & information-theoretic & 0.16 $\pm$ 0.10 & 85.40 & 0.05 & *   & 0.02 $\pm$ 0.07 & 236.25 & $<10^{-12}$\phantom{$<$} & *** \\ 
\cdashlinelr{1-10}
Granger & causal & 0.13 $\pm$ 0.09 & 57.40 & 1.1$\times 10^{3}$ & **  & $-$0.00 $\pm$ 0.05\phantom{$-$} & 1174.61 & $<10^{-12}$\phantom{$<$} & *** \\ 
\cdashlinelr{1-10}
Geometric & spatial & 0.09 $\pm$ 0.11 & 0.00 & 2.2$\times 10^{5}$ & *** & \textbf{0.32 $\pm$ 0.11} & 0.01 & 0.74 &     \\ 
\cdashlinelr{1-10}
\textbf{Heat-EV} & model-based & 0.16 $\pm$ 0.09 & 0.08 & 0.06 & {.}   & $-$0.05 $\pm$ 0.06\phantom{$-$} & 0.07 & $<10^{-12}$\phantom{$<$} & *** \\ 
\textbf{Heat-2} &  & 0.04 $\pm$ 0.10 & 0.06 & 2.9$\times 10^{9}$ & *** & $-$0.14 $\pm$ 0.09\phantom{$-$} & 0.07 & $<10^{-12}$\phantom{$<$} & *** \\ 
\textbf{Heat-2R} & & \textbf{0.20 $\pm$ 0.12} & 0.07 & { --} & & \underline{0.31 $\pm$ 0.09} & 0.08 & { --} & \\ 
\bottomrule
\end{tabular}
}
    \label{tab:fmri_alignment}
\end{table*}



\section{Discussion and Conclusion}
\label{sec:discussion_conclusion}

In this study, we introduced and evaluated a class of heat-based connectivity measures in both simulated and experimental settings. The first simulation demonstrated that 
the heat-based measures can capture undirected and directed graphs and improve structural fidelity. The subsequent simulations then showed
that the heat-based measures are less sensitive to artificial and spurious correlations, which we attribute to the model-based nature of these measures.

%
Experiment~I demonstrated that the regularised heat-based measure (Heat-2R) is overall the most effective measure for capturing informative discriminative features.
%
Experiment~II showed that the Heat-2R measure is also best at capturing modality-independent connectivity, which is further
%
 corroborated by a parameter sensitivity analysis in Supplemental Material~\ref{sm:parameter_sensitivity}.
However, on Dataset~IV, alignment between \gls{EEG} and \gls{fMRI} 
may be partly driven by spatial proximity rather than functional specificity, as suggested by the high alignment for geometric and correlation-based measures.

The proposed heat-based framework is suitable for settings with coarse temporal resolution, where signals can be approximated as smooth propagation over an interaction graph, and for settings where the primary objective is connectivity inference rather than detailed system modeling. In contrast, the framework may be less appropriate in regimes characterised by pronounced oscillatory dynamics, transient nonlinear events, or strong temporal effects such as delays, nonstationarity, or state-dependent switching.

The introduced measures have several limitations. 
Firstly, the simplicity of the underlying model may not adequately capture the complexity of \gls{EEG} data and can be sensitive to the sampling rate.
Secondly, none of the derived heat measures can determine the graph thermal diffusivity while simultaneously retaining relative structural fidelity.
Lastly, the heat measures rely on approximations to estimate the noise sources, since these sources are not directly measured.

Despite these limitations, we have demonstrated that the introduced heat-based connectivity measures provide fast, multivariate, scale-sensitive, directed, and robust connectivity estimation, making them well suited for graph-based applications, such as \glspl{GNN}.
%
Due to the simplicity of the model, our method may be applicable to a broader class of problems involving multivariate signals.

\section*{References}

\vspace*{-1.5em}
\bibliographystyle{IEEEtran}
\bibliography{bibliography}

\begin{thebibliography}{10}
\providecommand{\url}[1]{#1}
\csname url@samestyle\endcsname
\providecommand{\newblock}{\relax}
\providecommand{\bibinfo}[2]{#2}
\providecommand{\BIBentrySTDinterwordspacing}{\spaceskip=0pt\relax}
\providecommand{\BIBentryALTinterwordstretchfactor}{4}
\providecommand{\BIBentryALTinterwordspacing}{\spaceskip=\fontdimen2\font plus
\BIBentryALTinterwordstretchfactor\fontdimen3\font minus \fontdimen4\font\relax}
\providecommand{\BIBforeignlanguage}[2]{{%
\expandafter\ifx\csname l@#1\endcsname\relax
\typeout{** WARNING: IEEEtran.bst: No hyphenation pattern has been}%
\typeout{** loaded for the language `#1'. Using the pattern for}%
\typeout{** the default language instead.}%
\else
\language=\csname l@#1\endcsname
\fi
#2}}
\providecommand{\BIBdecl}{\relax}
\BIBdecl

\bibitem{wu2020comprehensive}
Z.~Wu, S.~Pan, F.~Chen, G.~Long, C.~Zhang, and P.~S. Yu, ``A comprehensive survey on graph neural networks,'' \emph{IEEE transactions on neural networks and learning systems}, vol.~32, no.~1, pp. 4--24, 2020.

\bibitem{thanou2017learning}
D.~Thanou, X.~Dong, D.~Kressner, and P.~Frossard, ``Learning heat diffusion graphs,'' \emph{IEEE Transactions on Signal and Information Processing over Networks}, vol.~3, no.~3, pp. 484--499, 2017.

\bibitem{goerttler2024understanding}
S.~Goerttler, M.~Wu, and F.~He, ``Understanding concepts in graph signal processing for neurophysiological signal analysis,'' in \emph{Machine Learning Applications in Medicine and Biology}.\hskip 1em plus 0.5em minus 0.4em\relax Springer, 2024, pp. 1--41.

\bibitem{klepl2024graph}
D.~Klepl, M.~Wu, and F.~He, ``Graph neural network-based eeg classification: A survey,'' \emph{IEEE Transactions on Neural Systems and Rehabilitation Engineering}, vol.~32, pp. 493--503, 2024.

\bibitem{stam2007phase}
C.~J. Stam, G.~Nolte, and A.~Daffertshofer, ``Phase lag index: assessment of functional connectivity from multi channel eeg and meg with diminished bias from common sources,'' \emph{Human brain mapping}, vol.~28, no.~11, pp. 1178--1193, 2007.

\bibitem{lachaux1999measuring}
J.-P. Lachaux, E.~Rodriguez, J.~Martinerie, and F.~J. Varela, ``Measuring phase synchrony in brain signals,'' \emph{Human brain mapping}, vol.~8, no.~4, pp. 194--208, 1999.

\bibitem{brookes2011investigating}
M.~J. Brookes, M.~Woolrich, H.~Luckhoo, D.~Price, J.~R. Hale, M.~C. Stephenson, G.~R. Barnes, S.~M. Smith, and P.~G. Morris, ``Investigating the electrophysiological basis of resting state networks using magnetoencephalography,'' \emph{Proceedings of the National Academy of Sciences}, vol. 108, no.~40, pp. 16\,783--16\,788, 2011.

\bibitem{bastos2016tutorial}
A.~M. Bastos and J.-M. Schoffelen, ``A tutorial review of functional connectivity analysis methods and their interpretational pitfalls,'' \emph{Frontiers in systems neuroscience}, vol.~9, p. 175, 2016.

\bibitem{mijalkov2022directed}
M.~Mijalkov, G.~Volpe, and J.~B. Pereira, ``Directed brain connectivity identifies widespread functional network abnormalities in parkinson’s disease,'' \emph{Cerebral cortex}, vol.~32, no.~3, pp. 593--607, 2022.

\bibitem{goerttler2024stochastic}
S.~Goerttler, F.~He, and M.~Wu, ``Stochastic graph heat modelling for diffusion-based connectivity retrieval,'' in \emph{2024 46th Annual International Conference of the IEEE Engineering in Medicine and Biology Society (EMBC)}, 2024, pp. 1--4.

\bibitem{atasoy2016human}
S.~Atasoy, I.~Donnelly, and J.~Pearson, ``Human brain networks function in connectome-specific harmonic waves,'' \emph{Nature communications}, vol.~7, no.~1, p. 10340, 2016.

\bibitem{guo2018functional}
D.~Guo, M.~Perc, T.~Liu, and D.~Yao, ``Functional importance of noise in neuronal information processing,'' \emph{Europhysics Letters}, vol. 124, no.~5, p. 50001, 2018.

\bibitem{penland1993prediction}
C.~Penland and T.~Magorian, ``Prediction of ni{\~n}o 3 sea surface temperatures using linear inverse modeling,'' \emph{Journal of Climate}, vol.~6, no.~6, pp. 1067--1076, 1993.

\bibitem{friston2003dynamic}
K.~J. Friston, L.~Harrison, and W.~Penny, ``Dynamic causal modelling,'' \emph{Neuroimage}, vol.~19, no.~4, pp. 1273--1302, 2003.

\bibitem{daunizeau2011dynamic}
J.~Daunizeau, O.~David, and K.~E. Stephan, ``Dynamic causal modelling: a critical review of the biophysical and statistical foundations,'' \emph{Neuroimage}, vol.~58, no.~2, pp. 312--322, 2011.

\bibitem{deco2011emerging}
G.~Deco, V.~K. Jirsa, and A.~R. McIntosh, ``Emerging concepts for the dynamical organization of resting-state activity in the brain,'' \emph{Nature reviews neuroscience}, vol.~12, no.~1, pp. 43--56, 2011.

\bibitem{breakspear2017dynamic}
M.~Breakspear, ``Dynamic models of large-scale brain activity,'' \emph{Nature neuroscience}, vol.~20, no.~3, pp. 340--352, 2017.

\bibitem{marimpis2021dyconnmap}
A.~D. Marimpis, S.~I. Dimitriadis, and R.~Goebel, ``Dyconnmap: Dynamic connectome mapping—a neuroimaging python module,'' \emph{Human Brain Mapping}, vol.~42, no.~15, pp. 4909--4939, 2021.

\bibitem{vinck2011improved}
M.~Vinck, R.~Oostenveld, M.~Van~Wingerden, F.~Battaglia, and C.~M. Pennartz, ``An improved index of phase-synchronization for electrophysiological data in the presence of volume-conduction, noise and sample-size bias,'' \emph{Neuroimage}, vol.~55, no.~4, pp. 1548--1565, 2011.

\bibitem{nolte2004identifying}
G.~Nolte, O.~Bai, L.~Wheaton, Z.~Mari, S.~Vorbach, and M.~Hallett, ``Identifying true brain interaction from eeg data using the imaginary part of coherency,'' \emph{Clinical neurophysiology}, vol. 115, no.~10, pp. 2292--2307, 2004.

\bibitem{kraskov2004estimating}
A.~Kraskov, H.~St{\"o}gbauer, and P.~Grassberger, ``Estimating mutual information,'' \emph{Physical Review E—Statistical, Nonlinear, and Soft Matter Physics}, vol.~69, no.~6, p. 066138, 2004.

\bibitem{ver2000non}
\BIBentryALTinterwordspacing
G.~Ver~Steeg, ``Non-parametric entropy estimation toolbox (npeet),'' \emph{Non-parametric entropy estimation toolbox (NPEET)}, 2000. [Online]. Available: \url{https://github.com/gregversteeg/NPEET}
\BIBentrySTDinterwordspacing

\bibitem{seabold2010statsmodels}
S.~Seabold and J.~Perktold, ``Statsmodels: econometric and statistical modeling with python.'' \emph{SciPy}, vol.~7, no.~1, pp. 92--96, 2010.

\bibitem{pedregosa2011scikit}
F.~Pedregosa, G.~Varoquaux, A.~Gramfort, V.~Michel, B.~Thirion, O.~Grisel, M.~Blondel, P.~Prettenhofer, R.~Weiss, V.~Dubourg \emph{et~al.}, ``Scikit-learn: Machine learning in python,'' \emph{the Journal of machine Learning research}, vol.~12, pp. 2825--2830, 2011.

\bibitem{python311}
\BIBentryALTinterwordspacing
P.~S. Foundation, ``Python 3.11.11,'' 2024. [Online]. Available: \url{https://www.python.org/downloads/release/python-31111/}
\BIBentrySTDinterwordspacing

\bibitem{singh2023evoked}
A.~Singh, R.~C. Cole, A.~I. Espinoza, J.~R. Wessel, J.~F. Cavanagh, and N.~S. Narayanan, ``Evoked mid-frontal activity predicts cognitive dysfunction in parkinson’s disease,'' \emph{Journal of Neurology, Neurosurgery \& Psychiatry}, vol.~94, no.~11, pp. 945--953, 2023.

\bibitem{miltiadous2023dataset}
A.~Miltiadous, K.~D. Tzimourta, T.~Afrantou, P.~Ioannidis, N.~Grigoriadis, D.~G. Tsalikakis, P.~Angelidis, M.~G. Tsipouras, E.~Glavas, N.~Giannakeas \emph{et~al.}, ``A dataset of eeg recordings from: Alzheimer's disease, frontotemporal dementia and healthy subjects,'' \emph{OpenNeuro.[Dataset]}, 2023.

\bibitem{gallego2023simultaneous}
J.~Gallego-Rudolf, M.~Corsi-Cabrera, L.~Concha, J.~Ricardo-Garcell, and E.~Pasaye-Alcaraz, ``Simultaneous and independent electroencephalography and magnetic resonance imaging: A multimodal neuroimaging dataset,'' \emph{Data in Brief}, vol.~51, p. 109661, 2023.

\bibitem{telesford2023open}
Q.~K. Telesford, E.~Gonzalez-Moreira, T.~Xu, Y.~Tian, S.~J. Colcombe, J.~Cloud, B.~E. Russ, A.~Falchier, M.~Nentwich, J.~Madsen \emph{et~al.}, ``An open-access dataset of naturalistic viewing using simultaneous eeg-fmri,'' \emph{Scientific Data}, vol.~10, no.~1, p. 554, 2023.

\bibitem{jenkinson2002improved}
M.~Jenkinson, P.~Bannister, M.~Brady, and S.~Smith, ``Improved optimization for the robust and accurate linear registration and motion correction of brain images,'' \emph{Neuroimage}, vol.~17, no.~2, pp. 825--841, 2002.

\bibitem{jenkinson2012fsl}
M.~Jenkinson, C.~F. Beckmann, T.~E. Behrens, M.~W. Woolrich, and S.~M. Smith, ``Fsl,'' \emph{Neuroimage}, vol.~62, no.~2, pp. 782--790, 2012.

\bibitem{fonov2011unbiased}
V.~Fonov, A.~C. Evans, K.~Botteron, C.~R. Almli, R.~C. McKinstry, D.~L. Collins, B.~D.~C. Group \emph{et~al.}, ``Unbiased average age-appropriate atlases for pediatric studies,'' \emph{Neuroimage}, vol.~54, no.~1, pp. 313--327, 2011.

\bibitem{avants2011reproducible}
B.~B. Avants, N.~J. Tustison, G.~Song, P.~A. Cook, A.~Klein, and J.~C. Gee, ``A reproducible evaluation of ants similarity metric performance in brain image registration,'' \emph{Neuroimage}, vol.~54, no.~3, pp. 2033--2044, 2011.

\bibitem{antspy}
{ANTsX Developers}, ``{ANTsPy: Advanced Normalization Tools in Python},'' \url{https://github.com/ANTsX/ANTsPy}, 2024, accessed: 2025-04-01.

\bibitem{gramfort2013meg}
A.~Gramfort, M.~Luessi, E.~Larson, D.~A. Engemann, D.~Strohmeier, C.~Brodbeck, R.~Goj, M.~Jas, T.~Brooks, L.~Parkkonen \emph{et~al.}, ``Meg and eeg data analysis with mne-python,'' \emph{Frontiers in Neuroinformatics}, vol.~7, p. 267, 2013.

\end{thebibliography}

\newpage
\begin{center}
    \large
    \textbf{Supplemental Material:} Connectivity Estimation using Stochastic Graph Heat Modelling
\end{center}

\section{Stochastic Graph Heat Equation}
\label{sec:sghm}

The heat equation is given by the following second-order differential equation:
\begin{align}
    \frac{\partial}{\partial t}x(s,t) = \Delta x(s,t)\label{eq:HE},
\end{align}
where $x(s, t)$ is a function, or field, in space and time and $\Delta$ is the Laplace operator. To model stochastic effects, a noise term can be added to Equation \eqref{eq:HE}, yielding the stochastic heat equation:
\begin{align}
    \frac{\partial}{\partial t}x(s,t) = \Delta x(s,t) + \sigma\frac{\partial}{\partial t} W(s,t)\label{eq:SHE},
\end{align}
where $W(s,t)$ and $\sigma$ denote a Wiener process and its scale, respectively.
Equation \eqref{eq:SHE} can be extended to spatially discretised fields, i.e., time-continuous multivariate signals $\mathbf{x}(t)$ with an underlying spatial structure, which can be algebraically represented by an adjacency matrix $\mathbf{A}$.
To discretise Equation \eqref{eq:SHE}, the continuous fields $x(s,t)$ and $W(s,t)$ are replaced by the spatial signal $\mathbf{x}(t)$ and a vector of Wiener processes $\mathbf{W}(t)=\left(W_1(t), ..., W_N(t)\right)^\top$, respectively. The Laplace operator $\Delta$ is replaced by the negative graph Laplacian $\mathbf{L}=\mathbf{D}-\mathbf{A}$ \cite{thanou2017learning}, where $\mathbf{D}\coloneqq \mathrm{diag}(\mathbf{A}\cdot \mathbf{1})$ is the degree matrix, yielding overall:
\begin{align}
    \frac{\partial}{\partial t}\mathbf{x}(t) = -\mathbf{L} \mathbf{x}(t) + \sigma\frac{\partial}{\partial t} \mathbf{W}(t).\label{eq:partial}
\end{align}

\subsection{Solution of the stochastic graph heat equation}
\label{ssec:stochastic_graph_heat}
An exact solution for the stochastic graph heat equation (Equation \eqref{eq:partial}) is given by:
\begin{align}
    \mathbf{x}(t) &= e^{-t\mathbf{L}}\left(\mathbf{x}_0 + \int_{t_0}^{t}e^{\tau\mathbf{L}}\sigma\frac{\partial}{\partial \tau}\mathbf{W}(\tau)d\tau\right), \label{eq:solution}
\end{align}
with the initial condition $\mathbf{x}(t_0) = \mathbf{x}_0$.
The validity of this solution can be shown by inserting $\mathbf{x}(t)$ into Equation \eqref{eq:partial}:
\begin{align}
    \frac{\partial}{\partial t} \mathbf{x}(t) = -&\mathbf{L} e^{-t\mathbf{L}}\left(\mathbf{x}_0 + \int_{t_0}^{t}e^{\tau\mathbf{L}}\sigma\frac{\partial}{\partial \tau}\mathbf{W}(\tau)d\tau\right)  \nonumber\\&+ e^{-t\mathbf{L}}e^{t\mathbf{L}}\sigma\frac{\partial}{\partial t}\mathbf{W}(t) \nonumber\\
    = -& \mathbf{L}\mathbf{x}(t) + \sigma \frac{\partial}{\partial t}\mathbf{W}(t).
\end{align}

\subsection{Model sampling and measurement noise}
Equation \eqref{eq:solution} describes the continuous evolution of the signal in time at discrete spatial locations. To model an experimental recording, the signal can be sampled at equidistant time steps $\Delta t$ determined by the device sampling rate.
Generally, the time step $\Delta t$ is sufficiently small, such that the signal evolution can be approximated as follows:
\begin{align}
    \mathbf{x}(t+\Delta t) =& e^{-(t+\Delta t)\mathbf{L}}\left(\mathbf{x}_0 + \int_{t_0}^{t+\Delta t}e^{\tau\mathbf{L}}\sigma\frac{\partial}{\partial \tau}\mathbf{W}(\tau)d\tau\right)  \nonumber\\=&e^{-\Delta t  \mathbf{L}}\left(\mathbf{x}(t) + \int_{t}^{t+\Delta t}e^{(\tau - t)\mathbf{L}}\sigma\frac{\partial}{\partial \tau}\mathbf{W}(\tau)d\tau\right) \nonumber\\
    \approx&e^{-\Delta t  \mathbf{L}}\left(\mathbf{x}(t) + \int_{t}^{t+\Delta t}\sigma\frac{\partial}{\partial \tau}\mathbf{W}(\tau)d\tau\right) \nonumber\\
    =&e^{-\Delta t  \mathbf{L}}\left(\mathbf{x}(t) + \boldsymbol{\epsilon}_\mathrm{int}\right),
\end{align}
where $\boldsymbol{\epsilon}_\mathrm{int} \sim \mathcal{N}(\mathbf{0}, \sigma_\mathrm{int}\Delta t)$ is a noise vector resulting from the definition of the Wiener process.

\section{Derivation of Undirected Positive-Definite Graphs}
\label{sec:der_ud}
To generate an undirected positive-definite graph, we begin by symmetrising a directed graph $\mathbf{A}^{\mathrm{d}}$:
\begin{align}
    \widetilde{A}_{ij}^{\mathrm{ud}}=\frac{A_{ij}^{\mathrm{d}} + {A_{ij}^{\mathrm{d}}}^\top}{2}.
\end{align}
The resulting matrix $\widetilde{\mathbf{A}}^{\mathrm{ud}}$ is not necessarily positive-definite; however, this is required by the correlation-based simulation described in Section \ref{ssec:correlation_simulation}. 
To enforce positive definiteness, we firstly compute the eigendecomposition:
\begin{align}
\widetilde{\mathbf{A}}^{\mathrm{ud}} &= \mathbf{Q} \widetilde{\boldsymbol{\Lambda}} \mathbf{Q}^\top,
\end{align}
where $\mathbf{Q}$ is the matrix of eigenvectors and $\widetilde{\boldsymbol{\Lambda}} = \operatorname{diag}(\tilde{\lambda}_1, \dots, \tilde{\lambda}_n)$ is the diagonal matrix of eigenvalues.
We then clip the eigenvalues to enforce a minimum value $\lambda_{\text{min}} = 0.01$:
\begin{align}
\hat{\lambda}_i &= \max(\tilde{\lambda}_i, \lambda_{\text{min}}),\\
\widehat{\boldsymbol{\Lambda}} &= \operatorname{diag}(\hat{\lambda}_1, \dots, \hat{\lambda}_n),
\end{align}
which allows us to reconstruct the adjacency matrix as a positive-definite matrix:
\begin{align}
\widehat{\mathbf{A}}^{\mathrm{ud}} &= \mathbf{Q} \widehat{\boldsymbol{\Lambda}} \mathbf{Q}^\top.
\end{align}
%
Lastly, to reset the diagonal elements to 1, we normalise $\widehat{\mathbf{A}}^{\mathrm{ud}}$ using the normalisation matrix $\mathbf{D}=\mathrm{diag}\left( 1/{\widehat{A}_{11}^{\mathrm{ud}}},\dots,1/{\widehat{A}_{NN}^{\mathrm{ud}}} \right)$:
\begin{align}
\mathbf{A}^{\mathrm{ud}} &= \mathbf{D}^{1/2}\widehat{\mathbf{A}}^{\mathrm{ud}}\mathbf{D}^{1/2}.
\end{align}

\section{Condition number control in regularisation}
\label{sm:regularisation}

We compute the condition number $\kappa$ of a matrix ${\mathbf{M}_0{\phantom{'}}}^{\hspace{-6pt}2} + \gamma \mathbbm{1}$, where $\gamma$ is defined as
\begin{align}
    \gamma \coloneqq \frac{- \sqrt{\kappa\left({\mathbf{M}_0{\phantom{'}}}^{\hspace{-6pt}2}\right)} \sigma_{\min}\left({\mathbf{M}_0{\phantom{'}}}^{\hspace{-6pt}2}\right) +  \sigma_{\max}\left({\mathbf{M}_0{\phantom{'}}}^{\hspace{-6pt}2}\right) }{ \sqrt{\kappa\left({\mathbf{M}_0{\phantom{'}}}^{\hspace{-6pt}2}\right)} - 1}.
\end{align}
We define $\sigma_{\max}^{(M)} \coloneqq \sigma_{\max}(\mathbf{M}_0^2)$ and $\lambda_{\min} \coloneqq \sigma_{\min}(\mathbf{M}_0^2)$ for brevity.
With this choice of $\gamma$, the condition number of the regularised matrix is given by  $\sqrt{\kappa\left({\mathbf{M}_0{\phantom{'}}}^{\hspace{-6pt}2}\right)}$:
\begin{figure*}[h]
\begin{align}
\kappa\left({\mathbf{M}_0{\phantom{'}}}^{\hspace{-6pt}2} + \gamma \mathbbm{1}\right) &= \frac{\sigma_{\max}\left({\mathbf{M}_0{\phantom{'}}}^{\hspace{-6pt}2} + \gamma \mathbbm{1}\right)}{\sigma_{\min}\left({\mathbf{M}_0{\phantom{'}}}^{\hspace{-6pt}2} + \gamma \mathbbm{1}\right)}= \frac{\sigma_{\max}\left({\mathbf{M}_0{\phantom{'}}}^{\hspace{-6pt}2}\right) + \sigma_{\max}\left(\gamma\mathbbm{1}\right)}{\sigma_{\min}\left({\mathbf{M}_0{\phantom{'}}}^{\hspace{-6pt}2}\right) + \sigma_{\min}\left(\gamma\mathbbm{1}\right)}\nonumber\\
&= \frac{\sigma_{\max}\left({\mathbf{M}_0{\phantom{'}}}^{\hspace{-6pt}2}\right)\left(\sqrt{\kappa\left({\mathbf{M}_0{\phantom{'}}}^{\hspace{-6pt}2}\right)} - 1\right) - \sqrt{\kappa\left({\mathbf{M}_0{\phantom{'}}}^{\hspace{-6pt}2}\right)} \sigma_{\min}\left({\mathbf{M}_0{\phantom{'}}}^{\hspace{-6pt}2}\right) +  \sigma_{\max}\left({\mathbf{M}_0{\phantom{'}}}^{\hspace{-6pt}2}\right)}{\sigma_{\min}\left({\mathbf{M}_0{\phantom{'}}}^{\hspace{-6pt}2}\right)\left(\sqrt{\kappa\left({\mathbf{M}_0{\phantom{'}}}^{\hspace{-6pt}2}\right)} - 1\right) - \sqrt{\kappa\left({\mathbf{M}_0{\phantom{'}}}^{\hspace{-6pt}2}\right)} \sigma_{\min}\left({\mathbf{M}_0{\phantom{'}}}^{\hspace{-6pt}2}\right) +  \sigma_{\max}\left({\mathbf{M}_0{\phantom{'}}}^{\hspace{-6pt}2}\right)}\nonumber\\
&= \frac{\sqrt{\kappa\left({\mathbf{M}_0{\phantom{'}}}^{\hspace{-6pt}2}\right)} \left(\sigma_{\max}\left({\mathbf{M}_0{\phantom{'}}}^{\hspace{-6pt}2}\right) - \sigma_{\min}\left({\mathbf{M}_0{\phantom{'}}}^{\hspace{-6pt}2}\right)\right)}{\sigma_{\max}\left({\mathbf{M}_0{\phantom{'}}}^{\hspace{-6pt}2}\right) - \sigma_{\min}\left({\mathbf{M}_0{\phantom{'}}}^{\hspace{-6pt}2}\right)}=\sqrt{\kappa\left({\mathbf{M}_0{\phantom{'}}}^{\hspace{-6pt}2}\right)}.
\end{align}
\end{figure*}

\section{Artificial Correlation-Based Simulation}
\label{ssec:correlation_simulation}
This section outlines the simulation of multivariate signals with an embedded correlation structure. 
The correlation structure is randomly generated as an undirected graph, as described in Section \ref{ssec:graph_structure}, and is incorporated into the signal in the frequency domain.

The first step of our simulation is to generate a random multivariate signal $\widetilde{\mathbf{X}}\in \mathbb{C}^{N \times N_t}$ in the frequency domain, with $N$ nodes and $N_t$ time samples:
\begin{align}
    \widetilde{X}_{ij} =
\begin{cases}
\sim \mathcal{NC}(0, 1) & \text{if } j <= N_\mathrm{max}, \\
0 & \text{else}.
\end{cases}
\end{align}
Here, $\mathcal{NC}$ is the standard complex normal distribution and $N_\mathrm{max}$ the number of frequency components.
$N_\mathrm{max}$ is linked to the cutoff frequency $f_\mathrm{max}$ as follows:
\begin{align}
    N_\mathrm{max}=\frac{f_\mathrm{max}}{f_s/2}N_t,
\end{align}
where $f_s$ denotes the sampling frequency. Note that $f_s/2$ is the Nyquist limit.

We use Cholesky decomposition to impose a correlation structure on $\widetilde{\mathbf{X}}$, which is a standard approach for controlling the covariance of samples drawn from a multivariate distribution\footnote{See J. E. Gentle, \textit{Computational statistics}. Springer, 2009, vol. 308.}. For a symmetric positive-definite matrix, the decomposition of the adjacency matrix $\mathbf{A}^\mathrm{ud}$ takes the form:
\begin{align}
    \mathbf{A}^\mathrm{ud}=\mathbf{L}\mathbf{L}^\top,
\end{align}
where $\mathbf{L}$ is a real lower triangular matrix with positive diagonal entries, not to be confused with the Laplacian matrix.

Subsequently, the Cholesky factor $\mathbf{L}$ is multiplied with the frequency signal $\widetilde{\mathbf{X}}$, yielding the modified frequency signal
\begin{align}
    \widetilde{\mathbf{X}}' = \mathbf{L}\widetilde{\mathbf{X}}.
\end{align}

Lastly, this modified signal is transformed to the temporal domain using the $N_t$-point inverse \gls{DFT} matrix $\mathbf{F}^{-1}$:
\begin{align}
    \mathbf{X} = \widetilde{\mathbf{X}}'\mathbf{F}^{-1}.
\end{align}
The expected value of $\mathbf{X}$ is zero, which follows from the linearity of the expectation value:
\begin{align}
    \mathbb{E}\left[\mathbf{X}\right] = \mathbb{E}\left[\mathbf{X}\right] = \mathbb{E}\left[ \mathbf{L}\widetilde{\mathbf{X}}\mathbf{F}^{-1} \right] = \mathbf{L}\,\mathbb{E}[\widetilde{\mathbf{X}}]\,\mathbf{F}^{-1}=0.
\end{align}
Furthermore, 
\begin{align}\mathbb{E}[\widetilde{\mathbf{X}}\widetilde{\mathbf{X}}^\mathrm{H}]=\frac{N_\mathrm{max}}{N_t}\cdot \mathbbm{1},\end{align}
since only $N_\mathrm{max}$ frequency components out of $N_t$ possible components are nonzero.

Importantly, the multivariate signal $\mathbf{X}$ embeds the correlation structure, as the following computation shows:
\begin{align}
    \mathrm{corr}(\mathbf{X},\mathbf{X})&=\mathbf{D}^{-1/2}\,\mathrm{cov}(\mathbf{X},\mathbf{X})\,\mathbf{D}^{-1/2}\nonumber\\&=\mathbf{D}^{-1/2}\,\mathbb{E}\left[ \mathbf{L}\widetilde{\mathbf{X}}\mathbf{F}^{-1}{\left(\mathbf{L}\widetilde{\mathbf{X}}\mathbf{F}^{-1}\right)}^\mathrm{H} \right]\,\mathbf{D}^{-1/2}\nonumber\\&=\mathbf{D}^{-1/2}\,\mathbb{E}\left[ \mathbf{L}\widetilde{\mathbf{X}}\mathbf{F}^{-1}{\mathbf{F}^{-1}}^\mathrm{H}\widetilde{\mathbf{X}}^\mathrm{H}{\mathbf{L}}^\top \right]\,\mathbf{D}^{-1/2}\nonumber\\
    &=\mathbf{D}^{-1/2}\,\left(\mathbf{L}\mathbf{L}^\top
    \mathbb{E}\left[ \widetilde{\mathbf{X}}\widetilde{\mathbf{X}}^\mathrm{H} \right]\frac{1}{N_t}
    \right)\,\mathbf{D}^{-1/2}\nonumber\\
    &=\mathbf{D}^{-1/2}\,\left(\mathbf{A}^\mathrm{ud}
    \frac{N_\mathrm{max}}{N_t}\frac{1}{N_t}
    \right)\,\mathbf{D}^{-1/2}\nonumber\\=\mathbf{A}^\mathrm{ud},\\
    \mathbf{D} &= \left(\mathbf{A}^\mathrm{ud}
    \frac{N_\mathrm{max}}{N_t^2}
    \right)\circ \mathbbm{1}=\frac{N_\mathrm{max}}{N_t^2}\mathbbm{1}.
\end{align}
Here, the operation $\circ$ denotes the Hadamard product, which is used to construct the diagonal variance matrix $\mathbf{D}$.
To account for external noise in this simulation, we further add normally distributed noise with standard deviation $\sigma$ to $\mathbf{X}$:
\begin{align}
    X_{ij}' = X_{ij} + \epsilon_{ij}, \quad \epsilon_{ij} \sim \mathcal{N}\left(0, \sigma^2\right).
\end{align}
This step is analogous to the incorporation of external noise in the heat-based simulation. Lastly, it should be noted that this step alters the correlation structure; however, the effect is minimal when $\sigma$ is small.

\subsection{Simulation results}
\begin{figure}[tbp]
    \centering
    \includegraphics[width=0.45\textwidth]{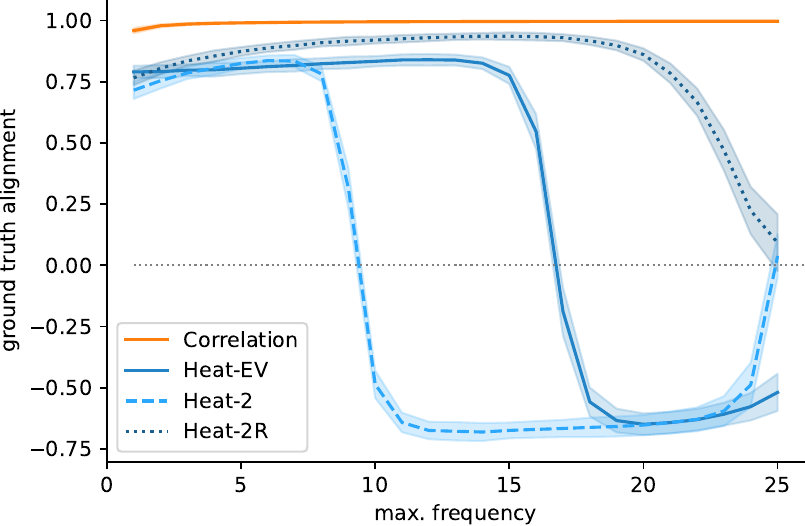}
    \caption[Ground truth alignment of the correlation-based simulation]{Mean ground truth alignment of the correlation-based simulation across maximum frequency values between 1 and 25. Shaded areas represent the standard error. The ground truth alignment for the correlation measure is close to 1, as intended by design. The heat connectivity measures are sensitive to the correlation structure when higher frequencies are excluded, but this sensitivity diminishes once sufficiently high frequencies are included.}
    \label{fig:sim_corr}
\end{figure}

Figure~\ref{fig:sim_corr} shows the ground truth alignment of the artificial correlation-based simulation for the heat-based connectivity measures and the correlation connectivity measure. As intended, the correlation connectivity measure is close to 1, despite the integration of external noise into the simulation. On the other hand, the ground truth alignment changes significantly with the maximum frequency. Generally, if only lower frequencies are included, the heat-based measures retrieve the correlation structure. Importantly, when higher frequencies are included, the heat-based measures become insensitive to the artificial correlation.

\section{Visual Results for Experiments I and II}
\begin{figure}[tbp]
\begin{minipage}[b]{0.48\textwidth}
    \vspace{-10pt}
    \hspace{4pt}
    \begin{tikzpicture}[baseline=(image.center)]
    \sffamily
          \node[anchor=south west, inner sep=0] (image) at (0,0) 
    {\includegraphics[width=0.88\textwidth,trim={5.5 -8.25 8 0},clip]{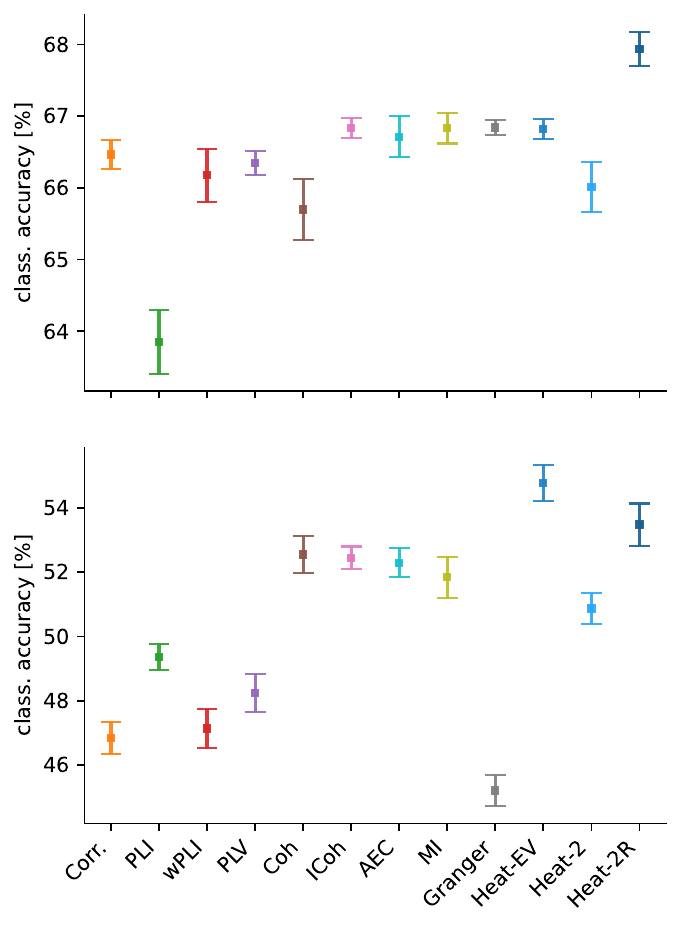}};
    \node[anchor=north west, xshift=-100pt + 123pt - 20pt, yshift=-500pt + 400pt + 106pt] 
        at (image.north west) {\large A}; 
    \node[anchor=north west, xshift=-100pt + 123pt - 20pt, yshift=-500pt + 300pt + 59pt] 
        at (image.north west) {\large B}; 
    \end{tikzpicture}
    \vspace{-11pt}
    \caption[\gls{SVM} classification accuracy over available connectivity measures]{\gls{SVM} classification accuracy over all available connectivity measures for Datasets~I (A) and II (B). The error bars denote the confidence intervals obtained from 100 bootstrap repetitions. Heat-2R (dark blue) achieves the highest accuracy on Dataset~I, while Heat-EV (blue) achieves the highest accuracy on Dataset~II.}
    \vspace{3.8pt}
    \label{fig:adj_classification}
\end{minipage}
\vfill
\begin{minipage}[b]{0.48\textwidth}
    
      \begin{tikzpicture}
      \sffamily
  \node[anchor=north west] at (0,3cm) {
    \includegraphics[width=0.9\textwidth,trim={2 240 0 -2.75},clip]{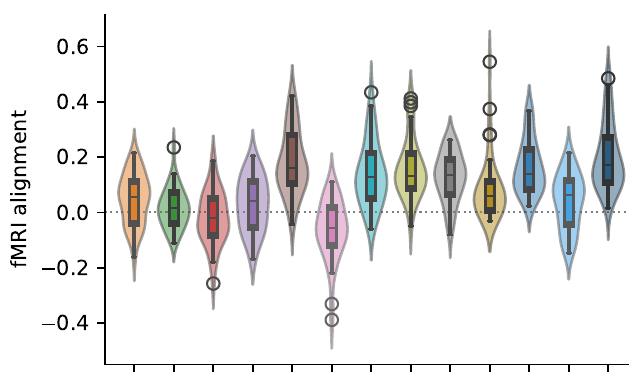}
  };
  \node[anchor=north west] at (0,-2cm) {
    \includegraphics[width=0.9\textwidth,trim={2 10 0 200},clip]{figures/violin_plot.pdf}
  };
    \node[anchor=west, xshift=-100pt + 123pt - 20pt + 12.5pt, yshift=0pt + 76pt] 
        at (0,0) {\large A}; 
    \node[anchor=west, xshift=-100pt + 123pt - 20pt + 12.5pt, yshift=-100pt+43pt] 
        at (0,0) {\large B}; 
      \end{tikzpicture}
    \vspace{-6pt}
    \caption[Violin plot showing the distribution of \gls{fMRI} alignment samples]{Violin plots of \gls{fMRI} alignment samples for Datasets~III (A) and IV (B). Heat-2R connectivity shows the highest alignment across both datasets. Overall ranking is largely consistent, with some differences between datasets for correlation, geometric, and heat measures.}
    \label{fig:violin}
\end{minipage}
\end{figure}

Figure~\ref{fig:adj_classification} presents the results of \textbf{Experiment I}, showing the \gls{SVM} classification accuracy for the three heat-based connectivity measures and nine baseline connectivity measures for Datasets I and II. The two heat-based connectivity measures Heat-EV and Heat-2R outperform all other measures on both datasets. While Heat-2R performs best on Dataset~I, Heat-EV achieves the highest accuracy on Dataset~II.

Figure~\ref{fig:violin} shows the results of \textbf{Experiment II} as violin plots of the alignment distribution across all participants for Datasets~III and IV. Heat-2R connectivity exhibits the highest \gls{fMRI} alignment overall, surpassed only by correlation and geometric connectivity on Dataset~IV. 
Additionally, its alignment distribution is mostly positive.
On Dataset~III, coherence connectivity follows Heat-2R in alignment.
Among the heat-based measures, Heat-EV ranks second across both datasets, with Heat-2 following in third.
Alignment is generally lower on Dataset~IV than on Dataset~III, except for correlation, geometric, and Heat-2R connectivity, which all show substantially higher alignment.
The measures \gls{PLI}, \gls{wPLI}, and \gls{iCoh} exhibit negative \gls{fMRI} alignment across both datasets.

\section{Parameter Sensitivity Analyses}
\label{sm:parameter_sensitivity}
\begin{figure*}[t!]
    \centering
    \begin{minipage}[b]{0.48\textwidth}
        \centering
            \begin{tikzpicture}[baseline=(image.center)]
    \sffamily
          \node[anchor=south west, inner sep=0] (image) at (0,0) 
        {\includegraphics[width=0.95\textwidth,trim={7 -2.75 7.5 0},clip]{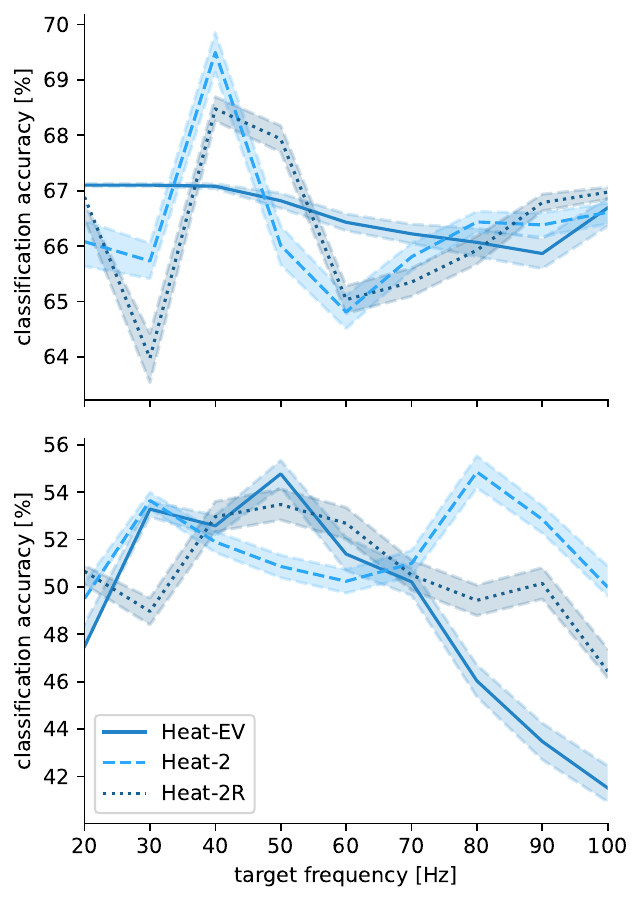}};

    \node[anchor=north west, xshift=-100pt + 120pt - 20pt, yshift=-500pt + 400pt + 107pt + 3pt] 
        at (image.north west) {\large A}; 
    \node[anchor=north west, xshift=-100pt + 120pt - 20pt, yshift=-500pt + 300pt + 44.5pt + 3pt] 
        at (image.north west) {\large B}; 
        \end{tikzpicture}
        \vspace{3pt}
        \caption[Parameter sensitivity analysis for connectivity-derived classification accuracy]{Parameter sensitivity analysis for connectivity-derived \gls{SVM} classification accuracy for Datasets I (A) and II (B). Performance is shown as bootstrapped mean with standard error across various target frequencies. Overall, performance is higher for intermediate sampling rate values. However, the performance is sensitive to the sampling rate, particularly for Dataset~I.}
        \label{fig:heat_param_sens_1}
    \end{minipage}\hfill
    \begin{minipage}[b]{0.48\textwidth}
        \centering
\begin{tikzpicture}

  \node[anchor=south west] at (0,6.2) {
    \includegraphics[width=0.95\textwidth,trim={7 221 7.5 -0},clip]{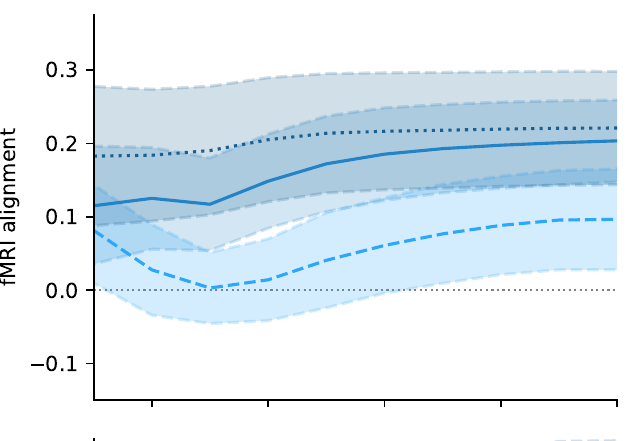}
  };

  \node[anchor=south west] at (0,0) { 
    \includegraphics[width=0.95\textwidth,trim={7 0 7.5 200},clip]{figures/parameter_sensitivity_alignment.pdf}
  };
\sffamily
    \node[anchor=north west, xshift=-300pt + 51pt + 280pt - 20pt, yshift=-500pt + 400pt + 106pt] 
        at (image.north west) {\large A}; 
    \node[anchor=north west, xshift=-300pt + 51pt +280pt - 20pt, yshift=-500pt + 300pt + 43.4pt] 
        at (image.north west) {\large B}; 
\end{tikzpicture}
        \vspace{3pt}
        \caption[Parameter sensitivity analysis for \gls{fMRI} alignment]{Parameter sensitivity analysis for \gls{fMRI} alignment for Datasets III (A) and IV (B). 
        Alignment is shown as bootstrapped mean with standard error across various target frequencies.
        Alignment decreases at lower sampling rates across both datasets. For sampling rates above 50\,Hz, alignment remains mostly stable, particularly for Heat-2R connectivity.}
        \label{fig:heat_param_sens_2}
    \end{minipage}
\end{figure*}

The key parameter governing heat-based connectivity retrieval is the sampling rate.
We perform parameter sensitivity analyses in Experiments I and II by varying the sampling rate from 20\,Hz to 100\,Hz.
Figure~\ref{fig:heat_param_sens_1} presents the parameter sensitivity analysis for \textbf{Experiment~I}, applied to Datasets I and II. On Dataset~I, the heat measures achieve peak performance at a sampling rate of 40\,Hz. On Dataset~II, the peak performances are achieved at 50\,Hz for Heat-EV and Heat-2R, and 80\,Hz for Heat-2. Generally, the performances tend to be higher for intermediate sampling rate values. The large fluctuations in the sensitivity analysis shows that the heat-based measures are sensitive to the sampling rate.

Figure~\ref{fig:heat_param_sens_2} shows the parameter sensitivity analysis for \textbf{Experiment~II}. The analysis shows that the \gls{fMRI} alignment tends to increases with increasing sampling rate. The parameter sensitivity analysis shows that the performance in terms of \gls{fMRI} alignment does not fluctuate much beyond a sampling rate of about 50\,Hz.

\end{document}